%% file: main_arxiv.tex
\documentclass[11pt]{article}
\usepackage[preprint]{acl}
\usepackage{times}
\usepackage{latexsym}
\usepackage[T1]{fontenc}
\usepackage[utf8]{inputenc}
\usepackage{microtype}
\usepackage{inconsolata}
\usepackage{graphicx}
\usepackage{booktabs}
\usepackage{multirow}
\usepackage{array}
\usepackage{tabularx}
\usepackage{enumitem}
\usepackage{algorithm}
\usepackage[noend]{algpseudocode}
\usepackage{colortbl}
\usepackage{pifont}
\usepackage{amsmath,amssymb}

\definecolor{groupbg}{HTML}{EFEFEF}
\definecolor{oursbg}{HTML}{E3EEF9}
\definecolor{upc}{HTML}{1B7F4B}
\definecolor{downc}{HTML}{C0392B}
\definecolor{flatc}{HTML}{7A7A7A}
\newcommand{\hd}[1]{\textbf{#1}}
\newcommand{\dup}[1]{\,{\scriptsize\textcolor{upc}{(#1)}}}
\newcommand{\ddn}[1]{\,{\scriptsize\textcolor{downc}{(#1)}}}
\newcommand{\dfl}[1]{\,{\scriptsize\textcolor{flatc}{(#1)}}}
\newcommand{\cmark}{\textcolor{upc}{\ding{51}}}
\newcommand{\xmark}{\textcolor{downc}{\ding{55}}}
\setlist[itemize]{itemsep=1pt,topsep=2pt,parsep=0pt,leftmargin=*}

\input{arxiv/numbers}

\newcommand{\grouprow}[2]{\rowcolor{groupbg}\multicolumn{#1}{c}{\textit{\textbf{#2}}\strut}}
\newcommand{\oursrow}{\rowcolor{oursbg}\bfseries}
\newlength{\fulltw}
\newlength{\fulltt}
\newcommand{\fulltable}[4]{%
  \begingroup
  \setlength{\tabcolsep}{0pt}%
  \settowidth{\fulltw}{\begin{tabular}{#3}#4\end{tabular}}%
  \setlength{\fulltt}{\dimexpr(#1-\fulltw)/(2*#2)\relax}%
  \setlength{\tabcolsep}{\fulltt}%
  \settowidth{\fulltw}{\begin{tabular}{#3}#4\end{tabular}}%
  \addtolength{\fulltt}{\dimexpr(#1-\fulltw)/(2*#2)\relax}%
  \setlength{\tabcolsep}{\fulltt}%
  \begin{tabular}{#3}#4\end{tabular}%
  \endgroup}
\newcommand{\aor}{AoR}
\newcommand{\supp}{\operatorname{Supp}}

\makeatletter
\ifacl@anonymize\else
\fi
\makeatother

\title{Accept or Repair: Preserving Passage-Supported Answers\\
in Contrastive Decoding}

\author{
  Jungseob Lee$^{1}$ \quad Seungyoon Lee$^{1}$ \quad Hyeonseok Moon$^{2}$ \\
  \bfseries Seongtae Hong$^{1}$ \quad Suhyune Son$^{1}$ \quad Sugyeong Eo$^{3}$ \\
  \bfseries Minhyuk Kim$^{1}$ \quad Chanjun Park$^{4\dagger}$ \quad Jaehyung Seo$^{5\dagger}$ \\
  {\normalfont\small $^{1}$Korea University \quad $^{2}$Sookmyung Women's University} \\
  {\normalfont\small $^{3}$Yonsei University Mirae Campus \quad $^{4}$Soongsil University} \\
  {\normalfont\small $^{5}$Konkuk University} \\
  {\normalfont\small\texttt{\{omanma1928,dltmddbs100,ghdchlwls123,ssh5131,mhkim0929\}@korea.ac.kr}} \\
  {\normalfont\small\texttt{hyns.moon@sookmyung.ac.kr} \quad \texttt{s.eo@yonsei.ac.kr}} \\
  {\normalfont\small\texttt{chanjun.park@ssu.ac.kr} \quad \texttt{seojae777@konkuk.ac.kr}} \\
  {\normalfont\small $^{\dagger}$Corresponding authors.}
}

\begin{document}
\maketitle

\input{arxiv/sections/abstract}

\input{arxiv/sections/intro}
\input{arxiv/sections/related}
\input{arxiv/sections/method}
\input{arxiv/sections/setup}
\input{arxiv/sections/results}
\input{arxiv/sections/analysis}
\input{arxiv/sections/conclusion}
\input{arxiv/sections/closing}

\bibliography{arxiv/references}

\appendix
\raggedbottom
\input{arxiv/sections/appendix}

\end{document}

%% file: arxiv/numbers.tex
\newcommand{\oursGainCF}{12.5}
\newcommand{\oursGainORD}{2.3}

\newcommand{\oursVsStrictCadCF}{2.9}
\newcommand{\oursVsStrictCadORD}{5.9}

\newcommand{\minGainModel}{2.9}
\newcommand{\oursCost}{1.71}

\newcommand{\taskWins}{44}
\newcommand{\taskCells}{45}
\newcommand{\cadBreakORD}{12.2}

\newcommand{\oursBreakCountORD}{4}
\newcommand{\rightCountORD}{24,639}

\newcommand{\accShareORDlo}{82}
\newcommand{\accShareORDhi}{86}

\newcommand{\baseRaiseLowHi}{1.1}
\newcommand{\oursRaiseLow}{2.3}

\newcommand{\plugMeanLo}{0.1}
\newcommand{\plugMeanHi}{2.7}

\newcommand{\plugCFAbsMax}{0.7}

\newcommand{\hoGainBase}{7.3}

\newcommand{\agGainBase}{1.2}

\newcommand{\olmoGainBest}{4.2}

\newcommand{\yiGainBest}{7.1}

\newcommand{\bigGainBest}{5.7}

\newcommand{\hugeGainBest}{4.8}

\newcommand{\oursVsSpanSwap}{3.8}
\newcommand{\oursVsSpanORD}{0.7}

\newcommand{\copyLossORDMag}{1.3}

\newcommand{\feedGainSwap}{1.6}
\newcommand{\feedGainORD}{0.2}
\newcommand{\oneCadGainSwap}{5.9}
\newcommand{\oneCadGainORD}{1.3}

\newcommand{\feedSame}{45}
\newcommand{\feedSup}{15}
\newcommand{\feedCadSup}{33}

\newcommand{\ablCicpMean}{0.4}

\newcommand{\spRepSwap}{67.6}
\newcommand{\spRepSwapOurs}{65.2}

\newcommand{\rtAccShareSwap}{70}

\newcommand{\scGainBestThirtyTwo}{5.7}

\newcommand{\scLossBestEight}{0.7}

\newcommand{\scLossBestThirtyTwo}{3.7}

\newcommand{\oursSwapGain}{8.3}

\newcommand{\noAcceptORDAbs}{75.7}

\newcommand{\strictResampleMeanAbs}{71.9}

\newcommand{\dcPSORD}{83.6}
\newcommand{\dcPlainSORD}{82.5}

\newcommand{\oursVsVetoMinCF}{1.2}
\newcommand{\oursVsVetoMinORD}{2.2}

\newcommand{\judgeCheckEM}{93}
\newcommand{\judgeCheckFzero}{8}
\newcommand{\judgeSwap}{3.9}

\newcommand{\judgeORD}{0.32}

\newcommand{\jgCheckEM}{100}
\newcommand{\jgCheckFzero}{9}

\newcommand{\wrapJudgeOrdLo}{0.5}
\newcommand{\wrapJudgeOrdHi}{5.7}

\newcommand{\wrapResidMax}{0.3}

\newcommand{\precSupLo}{72}
\newcommand{\precSupHi}{95}
\newcommand{\precUnsLo}{16}
\newcommand{\precUnsHi}{53}

\newcommand{\repPiLo}{50}
\newcommand{\repPiHi}{78}
\newcommand{\repURLo}{19}
\newcommand{\repURHi}{48}

\newcommand{\repTermLo}{12}
\newcommand{\repTermHi}{25}
\newcommand{\fbTermMaxMag}{2.2}

\newcommand{\lenGainLo}{0.8}
\newcommand{\lenGainHi}{9.1}
\newcommand{\lenHotTop}{1,060}
\newcommand{\accMemSwap}{1.7}
\newcommand{\gateRate}{20}
\newcommand{\gateMemRate}{12}

\newcommand{\gateSupMean}{3.6}

\newcommand{\gateOtherMeanHi}{2.2}

\newcommand{\gateMemSup}{99.5}
\newcommand{\gateMemOtherLo}{54}
\newcommand{\gateMemOtherHi}{55}

\newcommand{\gateJudgeMinMean}{0.6}
\newcommand{\gateJudgeSupQ}{1.1}
\newcommand{\gateJudgeOtherQ}{0.5}
\newcommand{\gateJudgeSupG}{2.3}
\newcommand{\gateJudgeOtherG}{1.8}
\newcommand{\latOursORD}{1.30}
\newcommand{\latCadORD}{1.32}
\newcommand{\resampleBrkORD}{2.7}
\newcommand{\memAgreeCad}{15.1}
\newcommand{\memDifferCad}{11.5}
\newcommand{\memAgreeStrict}{18.8}
\newcommand{\memDifferStrict}{15.1}
\newcommand{\memAgreeCoiecd}{14.7}
\newcommand{\memDifferCoiecd}{11.3}
\newcommand{\memAgreeId}{7.1}
\newcommand{\memDifferId}{9.1}
\newcommand{\memAgreeRpcd}{9.0}
\newcommand{\memDifferRpcd}{9.9}
\newcommand{\accPrecORD}{84.9}
\newcommand{\accRatioLo}{1.6}
\newcommand{\accRatioLoName}{NWCAD}
\newcommand{\accRatioHi}{4.9}
\newcommand{\accRatioHiName}{Strict CAD}
\newcommand{\nwcadCadMax}{0.5}
\newcommand{\strictResMinLead}{1.4}
\newcommand{\strictResSwapGain}{4.2}
\newcommand{\grdGainBaseMean}{7.3}
\newcommand{\gridMinMean}{72.5}
\newcommand{\gridMaxMean}{75.4}

\newcommand{\semSelfMinLead}{1.6}
\newcommand{\semNliMinLead}{2.0}
\newcommand{\rejOkQ}{86}
\newcommand{\rejOkG}{87}
\newcommand{\mPsgLo}{6.6}
\newcommand{\mPsgHi}{20.5}

\newcommand{\jbBrkCorOursORD}{1.6}
\newcommand{\jbBrkCorCadORD}{7.8}
\newcommand{\jgBrkCorOursORD}{1.7}
\newcommand{\jgBrkCorCadORD}{5.9}
\newcommand{\chgCadPct}{18.5}
\newcommand{\chgCadUnsup}{41}

\newcommand{\grEightMean}{74.4}
\newcommand{\grHalfMean}{73.1}

\newcommand{\grMemHalf}{4.4}
\newcommand{\rtOlmoGap}{4.2}
\newcommand{\rtYiGap}{5.6}
\newcommand{\rtBigGap}{3.5}
\newcommand{\joN}{167}
\newcommand{\joQwen}{0}
\newcommand{\joGemma}{1}

\newcommand{\joGoldQRej}{35}
\newcommand{\joGoldGRej}{0.5}

\newcommand{\ffMinLead}{1.6}
\newcommand{\ffMinName}{ID}
\newcommand{\jgSwap}{7.1}

\newcommand{\emSwapQ}{7.8}

\newcommand{\ablNoVetoFaithMag}{6.4}

\newcommand{\ablNoVetoCFMag}{3.2}

\newcommand{\ablKOneCMeanMag}{1.3}

\newcommand{\ablResampleMeanMag}{3.9}

\newcommand{\ablStrictResampleMeanMag}{3.0}

\newcommand{\dcCondCadORDMag}{4.1}
\newcommand{\dcCondStrictCadORDMag}{6.2}

\newcommand{\jgCadORDMag}{2.7}
\newcommand{\jgStrictCadORDMag}{4.6}

%% file: arxiv/sections/abstract.tex
\begin{abstract}
Large language models often answer from knowledge learned during training, even when a provided passage contradicts that knowledge.
To address this problem, contrastive decoding contrasts the predictions under the full and a modified prompt at every token, but it also alters correct answers that the passage supports.
We show that checking answers against the passage removes this trade-off.
Based on this observation, we propose Accept-or-Repair (AoR), a training-free procedure.
AoR keeps a sampled answer if the passage supports it.
Otherwise, it tries up to two replacements with contrastive decoding, returning one as soon as it passes this check.
It rejects replacements that repeat the answer generated without the passage when the passage favors another answer.
Our experiments show that AoR raises the score over sampling by \oursGainCF{} points on counterfactual passages and \oursGainORD{} on ordinary ones, surpassing eight contrastive and prompting baselines on both, including when each receives the same rejection rule.
Replacing only initial answers that the passage does not support improves the average score of every decoder.
These results suggest that models can follow passages that contradict their knowledge while preserving accuracy on passages that agree with it.
Code will be released at \url{https://github.com/js-lee-AI/AoR}.
\end{abstract}

%% file: arxiv/sections/intro.tex
\section{Introduction}
\label{sec:intro}

Large language models (LLMs) often prioritize knowledge learned during training over the information in a provided passage, even when the two conflict \citep{longpre-etal-2021-entity,xie2024chameleon}.
However, models should base their answers on the passage, whether it provides updated or private information or conflicts with their learned knowledge \citep{ram-etal-2023-context,zyskind-etal-2024-dont,shi-etal-2024-trusting,lee2025kullmrag}.

\begin{figure}[t]
\centering
\includegraphics[width=\columnwidth]{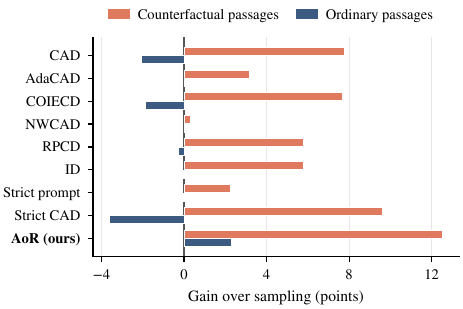}
\caption{Gain over plain sampling on counterfactual and ordinary passages, at the setting that each method chose on the development split. Test split, averaged over six models.}
\label{fig:teaser}
\end{figure}

To address this problem, contrastive decoding encourages models to follow the passage without additional training \citep{shi-etal-2024-trusting,wang-etal-2025-adacad}.
At each decoding step, it contrasts token predictions under the full and a modified prompt to favor information from the passage.
Context-aware decoding (CAD), for instance, removes the passage from the contrasting prompt \citep{shi-etal-2024-trusting}.
Methods also use signals from the model to adjust the contrast strength or decide when to apply it \citep{wang-etal-2025-adacad,yuan-etal-2024-discerning,tao2026noworse}.
These methods control answer generation without first checking whether a plain answer already follows the passage.
On ordinary passages, which agree with the model's parametric knowledge, CAD turns correct answers into errors more often than plain resampling.
The baselines that gain most on counterfactual passages reduce scores on ordinary passages, whereas the others leave ordinary scores nearly unchanged, as Figure~\ref{fig:teaser} shows.

For extractive questions, checking whether an answer occurs in the passage provides a cue for deciding whether to keep it.
On ordinary passages, most plain answers occur in the passage, and these answers are more accurate than those that do not.
We find that contrastive decoders lower the scores of these answers, with larger F1 losses for CAD when the model assigns the answer a higher probability without the passage.

\begin{figure*}[t]
\centering
\includegraphics[width=\textwidth]{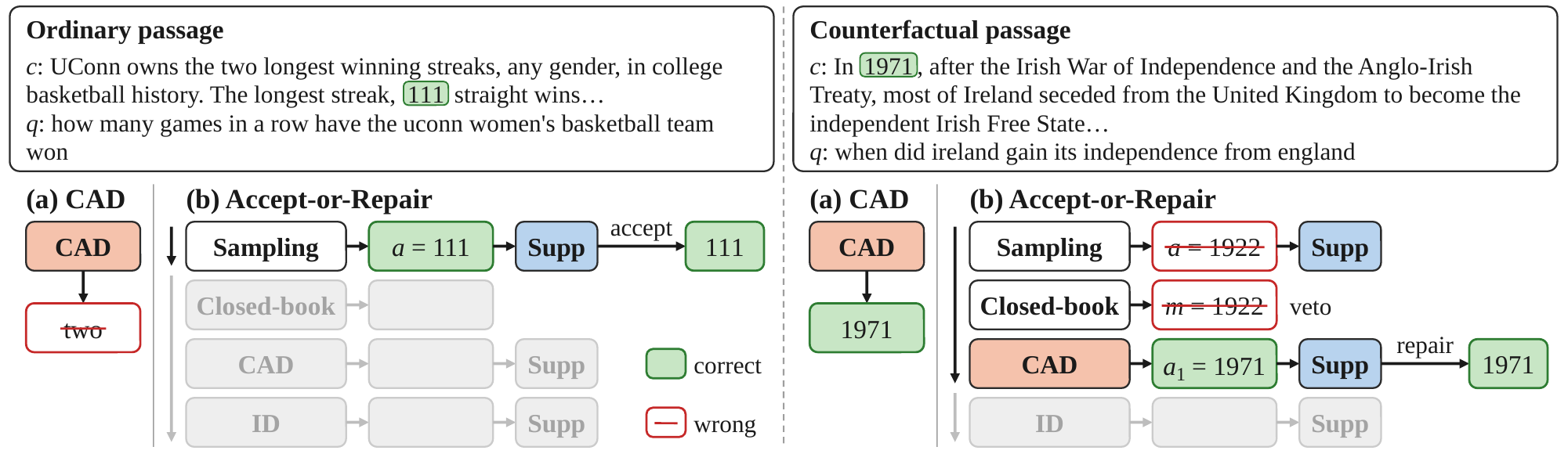}
\caption{Two test questions. (a) CAD contrasts every answer. (b) \aor{} accepts supported answers (Supp). When the passage favors another answer, the memory veto rejects repairs that repeat $m$, generated without the passage. Gray steps are skipped.}
\label{fig:method}
\end{figure*}

Based on this observation, we propose Accept-or-Repair (\aor{}), a training-free procedure that checks an initial answer before deciding whether to repair it.
\aor{} samples a plain answer and accepts it when the passage supports it.
For an unsupported initial answer, \aor{} draws an answer without the passage and uses it to screen replacements.
The memory veto of \aor{} rejects replacements that repeat this answer when the passage favors another answer.
\aor{} tries up to two replacements with contrastive decoding at the baselines' selected strengths and returns the first that passes both the support check and the veto.
Figure~\ref{fig:method} illustrates these decisions on two test questions.
CAD changes the correct answer 111 to \emph{two}, whereas \aor{} keeps 111 and replaces the memorized 1922 with 1971 from the passage.

Our experiments show that \aor{} improves average scores over plain sampling on both counterfactual and ordinary passages and outperforms contrastive and prompting baselines.
The lead remains when baselines receive the same memory veto, and LLM judges also rank \aor{} highest on average.

Replacing only unsupported plain answers improves the average score of each baseline.
The support test selects answers to repair more effectively than using the model's confidence, asking it to check its own answer or applying an entailment model.
These alternatives often accept answers from the model's memory even when the passage states a different answer.

We make three contributions.
\begin{itemize}
\item We show that contrastive decoders improve scores on unsupported initial answers but reduce scores on supported answers in ordinary passages, identifying a source of the trade-off between the two passage types.
\item We propose Accept-or-Repair, a training-free procedure that keeps supported answers and tries up to two contrastive repairs for the others, subject to the memory veto.
\item We show that \aor{} retains its lead on held-out benchmarks and across model families and sizes, and improves scores on both passage types when answers are given as one sentence.
\end{itemize}

%% file: arxiv/sections/related.tex
\section{Related Work}
\label{sec:related}

\paragraph{Knowledge conflicts.}
Models can favor memorized knowledge over conflicting passages and be misled by irrelevant context \citep{longpre-etal-2021-entity,xie2024chameleon,shi2023distracted,xu-etal-2024-knowledge-conflicts,lee2026madara}.
FaithEval and ConFiQA evaluate whether models base their answers on counterfactual passages \citep{ming2025faitheval,bi2025context}.
Prompting improves faithfulness to the passage through opinion-based questions and counterfactual demonstrations \citep{zhou-etal-2023-context}.

\paragraph{Decoding for context faithfulness.}
Contrastive decoding adjusts token scores by comparing predictions from the original input with those obtained without the passage, with an irrelevant passage, or with a modified instruction \citep{van-der-poel-etal-2022-mutual,shi-etal-2024-trusting,zhao-etal-2024-enhancing,kim2024instructive}.
Methods control contrastive decoding using distribution divergence, entropy, confidence signals, or a tunable factor \citep{wang-etal-2025-adacad,yuan-etal-2024-discerning,qiu-etal-2025-entropy,khandelwal-etal-2025-cocoa,kim-etal-2024-adaptive,li2026dccd,he2025cos}.
Gating and routing determine when to apply the contrast or which decoding regime to use \citep{tao2026noworse,jiang2026conflictaware,zhou2026dcrd,lee2026dart_routing,kang2026thinkfuse,lee2026redline}.
Other methods encourage models to follow the passage by adjusting token probabilities or steering hidden representations \citep{zhang2026cfb,gema-etal-2025-decore,zhao-etal-2025-steering,minder2025controllable,wang-etal-2025-continuously}.
\aor{} combines answer selection with contrastive decoding to keep supported initial answers and repair unsupported ones.

\paragraph{Answer selection and revision.}
Methods select among sampled answers by frequency or a trained verifier's score \citep{wang2023selfconsistency,cobbe2021gsm8k,lee2026directional,lee2026layermix}, and model cascades use learned estimates of answer quality to decide whether to query another model \citep{chen2024frugalgpt}.
Verification methods use self-critique, retrieved evidence, or attention signals to select or revise generated text \citep{asai2024selfrag,gao-etal-2023-rarr,dhuliawala-etal-2024-chain,chuang-etal-2024-lookback,lee2026pricecheck,lee2026agentloop}.
\aor{} selects initial answers for repair using a lexical test against the passage, without a trained verifier, self-verification prompts, or access to attention weights.

%% file: arxiv/sections/method.tex
\section{Accept-or-Repair}
\label{sec:method}

\paragraph{Setting.}
A model generates a short answer from a prompt containing a passage $c$ and a question $q$.
We write $a\sim D$ for an answer generated from this prompt using decoder $D$.
\aor{} keeps the initial answer from plain sampling if the passage supports it.
Otherwise, it generates replacements with contrastive decoders and checks them against the passage, as shown in Algorithm~\ref{alg:aor}.

\paragraph{Support test.}
For short free-form answers, the support test $\supp(a;c)$ accepts $a$ when its normalized form is nonempty and occurs in the normalized passage as a contiguous sequence of whole tokens.
Normalization lowercases the text, replaces ASCII punctuation with spaces, preserves non-ASCII punctuation and symbols as separate tokens, and removes articles.
For free-form questions, an answer that normalizes to yes or no passes the support test even if it does not occur in the passage.
For multiple-choice questions, the answer is an option letter. It passes the support test when the selected option's text occurs in the passage and no other option's text does.
For sentence answers, we exclude words that occur in the question and function words such as articles and prepositions. The answer passes the test when at least one word remains and all remaining words occur in the passage.

\begin{algorithm}[t]
\caption{Accept-or-Repair}
\label{alg:aor}
\begin{minipage}{\linewidth}
\begin{algorithmic}[1]
\Require passage $c$, question $q$, repair decoders $D_1,\dots,D_K$
\State $a \sim$ plain sampling
\If{$\supp(a;c)$} \Return $a$ \Comment{accept}
\EndIf
\State $m \sim$ plain sampling without the passage
\State $V \gets \{m\}$ if the passage favors another answer over $m$, else $\emptyset$
\For{$k=1,\dots,K$} \Comment{repair}
\State $a_k \sim D_k$, with the letter of $m$ masked if $m\in V$ on a multiple-choice question
\If{$\supp(a_k;c)$ and $a_k\notin V$} \Return $a_k$
\EndIf
\EndFor
\State \Return the first $a_k\notin V$ that states an answer, else $a$
\end{algorithmic}
\end{minipage}
\end{algorithm}

\paragraph{Accept.}
\aor{} first draws a plain answer $a$ and returns it unchanged when $\supp(a;c)$ holds, without applying contrastive decoding.

\paragraph{Repair.}
If the initial answer fails the support test, \aor{} draws candidate answers from $D_1,\dots,D_K$ in order.
It returns the first candidate that passes the test and is not rejected by the memory veto.
If no candidate passes both checks, \aor{} returns the first that states an answer and is not rejected by the veto, or the initial answer if no such candidate exists.
A candidate states an answer when its normalized text is nonempty for a free-form question or it gives a valid option letter for a multiple-choice question.
We use $K=2$, with CAD \citep{shi-etal-2024-trusting} followed by Instructive Decoding (ID) \citep{kim2024instructive}, and select the decoders and their order on the development split.
For ID, the contrasting prompt instructs the model to ignore the passage.
We write \aor{}($D$) for the variant that uses $D$ as its only repair without the memory veto.

\paragraph{Memory veto.}
For an unsupported initial answer, \aor{} draws a closed-book answer $m$ using plain sampling without the passage to screen repairs.
For free-form questions, if $m$ fails the support test, the veto rejects repairs that match $m$ after normalization.
Such repairs already fail the support test, so the veto prevents them from being returned as fallbacks.
For multiple-choice questions, an option scores 2 if its text occurs in the passage.
Otherwise, we exclude function words and use the fraction of the option's remaining words that occur in the passage.
If another option scores higher than the option selected by $m$, the veto masks the letter of $m$ during repair decoding.
For example, if $m$ is A and only option C's text occurs in the passage, the veto masks A and leaves B, C and D unmasked.
An option whose text occurs in the passage receives the maximum score, so the veto does not mask it.
On a paraphrased passage, the veto leaves $m$ unmasked when no other option scores higher.

\paragraph{Supported and unsupported answers.}
We analyze how retaining supported answers changes a decoder's score.
Let $S$ be the event that the initial plain answer $a$ passes the support test, and let $Q(a)$ and $Q(D)$ denote the scores in $[0,1]$ of $a$ and a draw from decoder $D$.
The expected score difference between $D$ and plain sampling is
\begin{equation}
\mathbb{E}[Q(D)-Q(a)]=P(S)\,\Delta_{S}+P(\neg S)\,\Delta_{\neg S},
\label{eq:split}
\end{equation}
where $\Delta_{S}=\mathbb{E}[Q(D)-Q(a)\mid S]$ and $\Delta_{\neg S}$ is the same difference on unsupported initial answers.
By retaining $a$ whenever $S$ holds, \aor{} preserves the score on supported answers and can change a correct initial answer only if that answer fails the support test.
For \aor{}($D$), which uses one repair without the memory veto, the output on unsupported initial answers is the draw from $D$, or $a$ if that draw states no answer.
With zero score for a draw that states no answer, the expected gain over $D$ is
\begin{equation}
\mathbb{E}[Q(\text{\aor{}}(D))-Q(D)]=-P(S)\,\Delta_{S}+\epsilon,
\label{eq:wrap}
\end{equation}
where $\epsilon\ge 0$ averages $Q(a)$ on these fallback cases and zero on all other cases.
Equation~\ref{eq:wrap} implies that \aor{}($D$) has a higher expected score than $D$ when $D$ lowers the average score on supported initial answers.

%% file: arxiv/sections/setup.tex
\section{Experimental Setup}
\label{sec:setup}

\input{arxiv/tables/main}

\paragraph{Tasks.}
We evaluate whether \aor{} follows counterfactual passages while preserving accuracy on ordinary passages.
In the main comparison, NQ-Swap \citep{longpre-etal-2021-entity} replaces the answer entity in a Natural Questions passage \citep{kwiatkowski-etal-2019-natural} with another entity of the same type, and FaithEval's counterfactual task \citep{ming2025faitheval} asks multiple-choice questions whose passage supports a counterfactual answer.
The ordinary tasks are NQ-Original, which uses the original passages and answers from NQ-Swap, SQuAD \citep{rajpurkar-etal-2016-squad} and HotpotQA \citep{yang-etal-2018-hotpotqa} in the distractor setting.
For HotpotQA, we concatenate two supporting paragraphs with up to eight distractors.
We also evaluate on the counterfactual subsets of ConFiQA \citep{bi2025context}.
For ordinary passages, we use ConFiQA's original passages and answers, TriviaQA \citep{joshi-etal-2017-triviaqa,fisch-etal-2019-mrqa} and SciQ \citep{welbl-etal-2017-crowdsourcing}.

\paragraph{Models.}
The main comparison uses Qwen3 models with 1.7B, 4B, 8B and 14B parameters \citep{yang2025qwen3}, Llama-3.1-8B-Instruct \citep{grattafiori2024llama} and Mistral-7B-Instruct-v0.3 \citep{jiang2023mistral,mistral2024instructv03}.
We also evaluate Qwen3-32B, Llama-3.1-70B-Instruct, OLMo-2-13B-Instruct \citep{olmo2024olmo2} and Yi-1.5-9B-Chat \citep{young2024yi,01ai2024yi15chat} to examine performance across model sizes and families.
All models are instruction-tuned, and Qwen3 runs in non-thinking mode.
For each model, the methods share the temperature and top-$p$ sampling settings \citep{holtzman2020curious} listed in Table~\ref{tab:sampling}, with top-$k$ filtering enabled for Qwen3.
Temperature scaling and filtering follow any contrastive adjustment of token scores.
We share sampling noise across methods within each run and average the scores over three seeds.

\paragraph{Baselines.}
CAD, AdaCAD, COIECD and No-Worse CAD (NWCAD) use the question alone as the contrastive input \citep{shi-etal-2024-trusting,wang-etal-2025-adacad,yuan-etal-2024-discerning,tao2026noworse}.
Random-passage CD (RPCD) contrasts with another example's passage \citep{zhao-etal-2024-enhancing}, and ID with an instruction to ignore the passage \citep{kim2024instructive}.
To compare their contrastive inputs, we use the CAD scoring rule for both baselines.
Strict prompt asks the model to answer only from the passage \citep{zhou-etal-2023-context}, and Strict CAD applies CAD to that prompt.
Contrastive settings are selected on the development split and kept fixed across models and tasks.
We choose the setting with the highest average gain over sampling across the two task groups, retaining a recommended setting when it is within 0.1 points of the best.
Strict CAD uses CAD's selected strength, and \aor{} uses the selected CAD and ID strengths for its repairs.

\paragraph{Evaluation.}
We use token-level F1 for free-form answers and accuracy for multiple-choice answers.
For the main comparison, we average task scores within the counterfactual and ordinary groups and report Mean as the average of the two group scores.
We assess score differences with 95\% confidence intervals from paired bootstrap resampling \citep{berg-kirkpatrick-etal-2012-empirical}.
Cost is the number of tokens processed across decoding calls, including prompts, relative to plain sampling.
Appendices~\ref{app:prompts} through~\ref{app:strengths} give the prompts, settings and scoring procedures.

%% file: arxiv/tables/main.tex
% generated by scripts/v8_numbers.py from results/v7/analysis_20261006_r5/out/aor_table_test_r5.json; split=test
\begin{table*}[t]
\centering\small
\fulltable{\textwidth}{10}{lccccccccc}{%
\toprule
 & \multicolumn{3}{c}{\hd{Counterfactual}} & \multicolumn{4}{c}{\hd{Ordinary}} & & \\
\cmidrule(lr){2-4}\cmidrule(lr){5-8}
\hd{Method} & \hd{NQ-Swap} & \hd{FaithEval} & \hd{Avg.} & \hd{NQ-Orig.} & \hd{SQuAD} & \hd{HotpotQA} & \hd{Avg.} & \hd{Mean} & \hd{Cost} \\
\midrule
Sampling & 56.9 & 63.6 & 60.3 & \underline{76.0} & 84.5 & 63.5 & 74.7 & 67.5 & 1.00 \\
CAD {\scriptsize\citep{shi-etal-2024-trusting}} & 62.4 & 73.7 & 68.1\dup{+7.8} & 72.5 & 84.6 & 60.8 & 72.6\ddn{$-$2.1} & 70.3 & 1.18 \\
AdaCAD {\scriptsize\citep{wang-etal-2025-adacad}} & 61.4 & 65.5 & 63.4\dup{+3.2} & 75.6 & 85.5 & 63.2 & \underline{74.8}\dfl{+0.1} & 69.1 & 1.18 \\
COIECD {\scriptsize\citep{yuan-etal-2024-discerning}} & 62.2 & 73.7 & 68.0\dup{+7.7} & 72.7 & 84.5 & 61.1 & 72.8\ddn{$-$1.9} & 70.4 & 1.18 \\
NWCAD {\scriptsize\citep{tao2026noworse}} & 57.3 & 63.8 & 60.6\dup{+0.3} & \underline{76.0} & 84.5 & \underline{63.7} & \underline{74.8}\dfl{+0.1} & 67.7 & 1.18 \\
RPCD {\scriptsize\citep{zhao-etal-2024-enhancing}} & 61.0 & 71.1 & 66.1\dup{+5.8} & 75.4 & 85.3 & 62.5 & 74.4\dfl{$-$0.3} & 70.2 & 2.00 \\
ID {\scriptsize\citep{kim2024instructive}} & 64.3 & 67.8 & 66.0\dup{+5.8} & 74.8 & \underline{85.6} & 63.6 & 74.7\dfl{0.0} & 70.4 & 2.01 \\
Strict prompt {\scriptsize\citep{zhou-etal-2023-context}} & 60.3 & 64.8 & 62.5\dup{+2.2} & \underline{76.0} & 84.9 & 63.1 & 74.6\dfl{0.0} & 68.6 & 1.03 \\
\begin{tabular}[t]{@{}l@{}}Strict CAD\\{\scriptsize\citep{zhou-etal-2023-context,shi-etal-2024-trusting}}\end{tabular} & \underline{64.7} & \underline{75.1} & \underline{69.9}\dup{+9.6} & 70.9 & 84.1 & 58.3 & 71.1\ddn{$-$3.6} & \underline{70.5} & 1.21 \\
\oursrow \aor{} (ours) & \textbf{65.2} & \textbf{80.4} & \textbf{72.8}\dup{+12.5} & \textbf{78.2} & \textbf{87.5} & \textbf{65.2} & \textbf{77.0}\dup{+2.3} & \textbf{74.9} & 1.71 \\
\bottomrule
}
\caption{Test scores averaged over six models: F1 for free-form tasks, accuracy for FaithEval. Mean averages both groups. Parentheses show differences from sampling, colored when their 95\% intervals exclude zero. Cost is the token count relative to sampling.}
\label{tab:main}
\end{table*}

%% file: arxiv/sections/results.tex
\begin{figure*}[t]
\centering
\includegraphics[width=\textwidth]{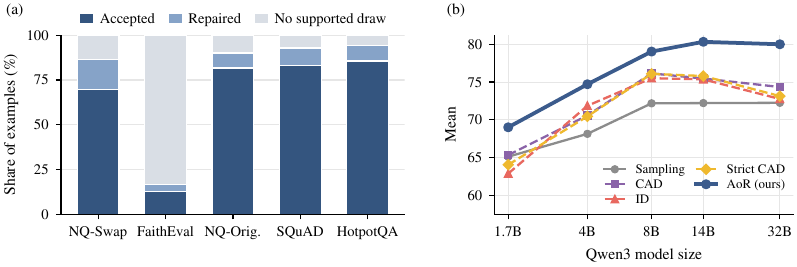}
\caption{(a) How \aor{} ends on each task over six models: the plain answer is accepted, a repair is supported, or no draw is supported. (b) Mean on Qwen3 models of five sizes.}
\label{fig:analysis}
\end{figure*}

\section{Results}
\label{sec:results}

\paragraph{Main results.}
We find that \aor{} is the only method to improve significantly over sampling in both task groups, by \oursGainCF{} points on counterfactual passages and \oursGainORD{} on ordinary passages, as Table~\ref{tab:main} shows.
It scores highest on each task and significantly outperforms the baselines in Mean.
Strict CAD, CAD and COIECD make the largest counterfactual gains among the baselines but lose on ordinary passages, whereas the other baselines leave ordinary scores nearly unchanged and gain less on counterfactual passages.
The lead in Mean holds on each model, as reported in Appendix~\ref{app:models}.
\aor{} has a token cost of \oursCost{} times that of sampling, lower than RPCD and ID.

\input{arxiv/tables/anatomy}
\input{arxiv/tables/plugin}

\begin{figure}[t]
\centering
\includegraphics[width=\columnwidth]{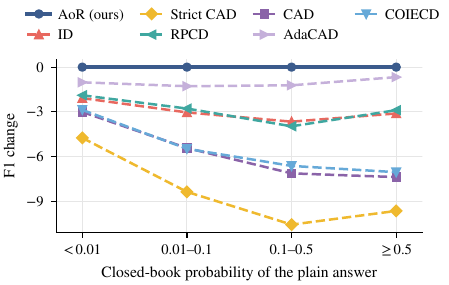}
\caption{F1 change on the supported plain answers of the ordinary tasks, binned by the probability that the model gives the same answer without the passage.}
\label{fig:mech}
\end{figure}

\paragraph{Where the gains come from.}
We find that the baselines improve on unsupported initial answers in both groups but lose on supported answers in ordinary passages, as Table~\ref{tab:anatomy} shows.
Figure~\ref{fig:mech} shows larger F1 losses for Strict CAD, CAD and COIECD in bins where the model assigns higher probability to the initial answer without the passage.
By retaining supported answers, \aor{} avoids this loss term in Equation~\ref{eq:split}.
It turns \oursBreakCountORD{} of the \rightCountORD{} correct initial answers on ordinary passages into errors and corrects more wrong initial answers on counterfactual passages than the baselines.
These findings suggest that retaining supported answers helps turn contrastive repairs into gains on ordinary passages.

\begin{figure*}[t]
\centering
\includegraphics[width=\textwidth]{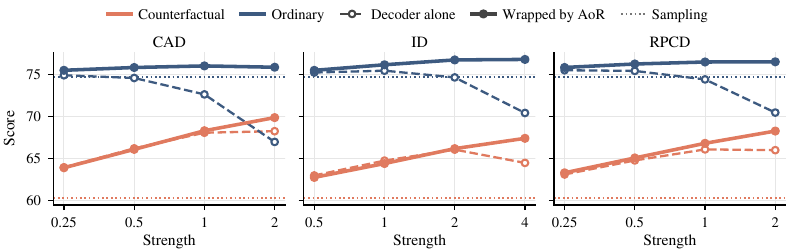}
\caption{Counterfactual and ordinary averages of CAD, ID and RPCD at four strengths, alone and wrapped by \aor{} as the only repair. Dotted lines mark sampling.}
\label{fig:strength}
\end{figure*}

\paragraph{Wrapping other decoders.}
We find that replacing only unsupported initial answers significantly improves the Mean of each baseline, as Table~\ref{tab:plugin} shows.
Each baseline serves as \aor{}'s only repair in this comparison, without the memory veto.
On ordinary passages, the gains recover the losses on supported answers in Table~\ref{tab:anatomy}, as Equation~\ref{eq:wrap} predicts.
Counterfactual scores remain nearly unchanged.

We also compare four contrast strengths for CAD, ID and RPCD to test whether this benefit depends on the selected setting.
Without \aor{}, all three decoders score below sampling on ordinary passages at the strongest setting, as Figure~\ref{fig:strength} shows.
With \aor{}, counterfactual scores rise with contrast strength while ordinary scores stay above sampling, giving each decoder its highest Mean at the strongest setting.
These findings suggest that retaining supported answers allows stronger contrastive repairs while preserving performance on ordinary passages.

\input{arxiv/tables/ablation}
\input{arxiv/tables/gate}

\paragraph{Ablations.}
Keeping supported initial answers improves ordinary scores and reduces token cost, as Table~\ref{tab:ablation} shows.
Allowing partial matches between answers and passage spans lowers the Mean and admits more memorized answers on NQ-Swap.
Removing the memory veto lowers FaithEval accuracy, and removing either repair lowers scores on both passage types.

Replacing contrastive repairs with samples from the strict or plain prompt lowers \aor{}'s Mean by \ablStrictResampleMeanMag{} to \ablResampleMeanMag{} points, with larger losses on counterfactual passages.
The strict-prompt variant still outperforms the baselines in Mean.
Majority voting \citep{wang2023selfconsistency} yields lower scores than \aor{} on both passage types and uses more tokens.
These comparisons suggest that \aor{} benefits from both checking answers against the passage and using contrastive decoding to generate repairs.

Adding Strict CAD restricted to passage spans as a third repair raises the Mean at a higher token cost.
The lead of \aor{} over the baselines in Mean also holds across the 16 combinations of CAD and ID strengths, as Appendix~\ref{app:further} shows.

\paragraph{Choosing what to repair.}
The support test gives the largest gains on NQ-Swap and ordinary passages among the selection criteria in Table~\ref{tab:gate}.
We compare it with answer probability, the probability gain from adding the passage, a model self-check \citep{kadavath2022language} and an entailment score.
To isolate the selection criterion, each alternative replaces the same number of initial answers as the support test, using the same CAD outputs without further checks.

On NQ-Swap, the support test selects \gateMemSup{}\% of initial answers that repeat the answer from the original passage, compared with \gateMemOtherLo{}--\gateMemOtherHi{}\% for answer probability, self-check and entailment.
This difference suggests that passage support helps identify memorized answers that these three criteria miss.

%% file: arxiv/tables/anatomy.tex
% generated by scripts/v8_numbers.py (restructure 2026-10-08) from break_fix_test_r5.json, review_c3/lowered.json, gain_anatomy_test_r5.json and review_c2/decomp.json; split=test
\begin{table*}[t]
\centering\small
\fulltable{\textwidth}{10}{lccccccccc}{%
\toprule
 & \multicolumn{3}{c}{\hd{Counterfactual}} & \multicolumn{4}{c}{\hd{Ordinary}} & \multicolumn{2}{c}{\hd{Repaired part}} \\
\cmidrule(lr){2-4}\cmidrule(lr){5-8}\cmidrule(lr){9-10}
\hd{Method} & $S$ & $\neg S$ & \hd{Fixed (\%)} & $S$ & $\neg S$ & \hd{Broken (\%)} & \hd{Lowered (\%)} & \hd{CF} & \hd{Ord.} \\
\midrule
Sampling & -- & -- & -- & -- & -- & -- & -- & 37.2 & 35.5 \\
CAD & $-$0.2 & +8.0 & 22.0 & $-$3.4 & +1.3 & 12.2 & 12.9 & 52.7 & 43.7 \\
AdaCAD & \textbf{+0.3} & +2.9 & 9.4 & $-$0.9 & +0.9 & 3.9 & 5.3 & 44.7 & 41.1 \\
COIECD & $-$0.2 & +7.9 & 21.6 & $-$3.3 & +1.4 & 11.9 & 12.6 & 52.5 & 43.9 \\
NWCAD & $-$0.1 & +0.4 & 1.2 & \underline{$-$0.1} & +0.2 & \underline{0.9} & \textbf{1.4} & 38.2 & 36.6 \\
RPCD & $-$0.7 & +6.5 & 17.6 & $-$2.1 & +1.8 & 9.7 & 10.1 & 50.6 & 46.3 \\
ID & \underline{0.0} & +5.8 & 19.2 & $-$2.1 & \underline{+2.1} & 8.7 & 9.4 & 51.6 & \underline{48.0} \\
Strict prompt & $-$0.2 & +2.5 & 9.3 & $-$0.9 & +0.8 & 3.7 & 4.9 & 43.9 & 40.4 \\
Strict CAD & \underline{0.0} & \underline{+9.6} & \underline{28.0} & $-$5.2 & +1.6 & 15.8 & 15.9 & \underline{56.5} & 45.1 \\
\oursrow \aor{} (ours) & \underline{0.0} & \textbf{+12.5} & \textbf{31.2} & \textbf{0.0} & \textbf{+2.3} & \textbf{0.0} & \underline{3.3} & \textbf{60.5} & \textbf{49.5} \\
\bottomrule
}
\caption{Where the gains come from. $S$, $\neg S$: share-weighted difference to sampling on supported and unsupported plain answers; Fixed, Broken: wrong plain answers made right, correct ones made wrong; Lowered: F1 below the plain answer; Repaired part: scores where \aor{} repairs.}
\label{tab:anatomy}
\end{table*}

%% file: arxiv/tables/plugin.tex
% generated by scripts/v8_numbers.py from plugin_ci_test_r5.json; split=test
\begin{table}[t]
\centering\small
\fulltable{\columnwidth}{4}{lccc}{%
\toprule
\hd{Wrapped decoder} & \hd{CF} & \hd{Ord.} & \hd{Mean} \\
\midrule
CAD & 68.3 & 76.0\dup{+3.4} & 72.2\dup{+1.8} \\
AdaCAD & 63.2 & 75.6\dup{+0.9} & 69.4\dup{+0.3} \\
COIECD & 68.2 & 76.1\dup{+3.3} & 72.1\dup{+1.7} \\
NWCAD & 60.7 & 74.9\dup{+0.1} & 67.8\dup{+0.1} \\
RPCD & 66.8 & 76.5\dup{+2.1} & 71.6\dup{+1.4} \\
ID & 66.2 & 76.7\dup{+2.1} & 71.5\dup{+1.1} \\
Strict prompt & 63.0 & 75.5\dup{+0.9} & 69.3\dup{+0.7} \\
Strict CAD & 70.1 & 76.3\dup{+5.2} & 73.2\dup{+2.7} \\
\bottomrule
}
\caption{Each decoder $D$ wrapped by \aor{} as its only repair, without the veto. On ordinary passages and in Mean, parentheses give the difference to $D$ alone, colored when its 95\% interval excludes zero.}
\label{tab:plugin}
\end{table}

%% file: arxiv/tables/ablation.tex
% generated by scripts/v8_numbers.py from ablate_ci_test_r5.json; split=test
\begin{table}[t]
\centering\small
\fulltable{\columnwidth}{5}{lcccc}{%
\toprule
\hd{Variant} & \hd{CF} & \hd{Ord.} & \hd{Mean} & \hd{Cost} \\
\midrule
\oursrow \aor{} & 72.8 & 77.0 & 74.9 & 1.71 \\
\grouprow{5}{Components} \\
Repair every answer & \textcolor{upc}{+0.7} & \textcolor{downc}{$-$1.3} & \textcolor{downc}{$-$0.3} & 2.71 \\
No memory veto & \textcolor{downc}{$-$3.2} & \textcolor{flatc}{0.0} & \textcolor{downc}{$-$1.6} & 1.63 \\
Any occurring option supported & \textcolor{downc}{$-$0.9} & \textcolor{flatc}{0.0} & \textcolor{downc}{$-$0.5} & 1.65 \\
\grouprow{5}{Repairs} \\
CAD only & \textcolor{downc}{$-$1.6} & \textcolor{downc}{$-$1.0} & \textcolor{downc}{$-$1.3} & 1.46 \\
ID only & \textcolor{downc}{$-$2.4} & \textcolor{downc}{$-$0.3} & \textcolor{downc}{$-$1.3} & 1.73 \\
ID, then CAD & \textcolor{downc}{$-$1.2} & \textcolor{upc}{+0.1} & \textcolor{downc}{$-$0.6} & 1.86 \\
CAD, then Strict CAD & \textcolor{downc}{$-$0.4} & \textcolor{downc}{$-$0.5} & \textcolor{downc}{$-$0.4} & 1.61 \\
CAD, ID, then CAD & \textcolor{upc}{+0.2} & \textcolor{upc}{+0.1} & \textcolor{upc}{+0.1} & 1.81 \\
CAD, ID, then span Strict CAD & \textcolor{upc}{+1.4} & \textcolor{upc}{+0.9} & \textcolor{upc}{+1.2} & 1.81 \\
\grouprow{5}{Repairs without contrast} \\
Resample & \textcolor{downc}{$-$6.1} & \textcolor{downc}{$-$1.6} & \textcolor{downc}{$-$3.9} & 1.56 \\
Resample, strict prompt & \textcolor{downc}{$-$4.9} & \textcolor{downc}{$-$1.2} & \textcolor{downc}{$-$3.0} & 1.56 \\
\grouprow{5}{Majority votes} \\
Three plain answers & \textcolor{downc}{$-$12.4} & \textcolor{downc}{$-$2.1} & \textcolor{downc}{$-$7.2} & 3.00 \\
Plain, CAD and ID & \textcolor{downc}{$-$7.8} & \textcolor{downc}{$-$1.6} & \textcolor{downc}{$-$4.7} & 4.19 \\
\bottomrule
}
\caption{Variants of \aor{} and majority votes. Score differences are relative to \aor{}; color indicates a 95\% interval excluding zero. Cost is the token count relative to sampling.}
\label{tab:ablation}
\end{table}

%% file: arxiv/tables/gate.tex
% generated by scripts/v8_numbers.py from review_c3/matched_gate.json; split=test
\begin{table}[t]
\centering\small
\fulltable{\columnwidth}{4}{lccc}{%
\toprule
\hd{Gate} & \hd{NQ-Swap} & \hd{Ord.} & \hd{Mean} \\
\midrule
CAD on every answer & \textcolor{upc}{+5.5} & \textcolor{downc}{$-$2.1} & \textcolor{upc}{+1.7} \\
Same as closed-book answer & \textcolor{upc}{+1.5} & \textcolor{downc}{$-$0.8} & \textcolor{upc}{+0.3} \\
Answer probability & \textcolor{upc}{+4.2} & \textcolor{upc}{+0.2} & \textcolor{upc}{+2.2} \\
Probability gain from passage & \textcolor{upc}{+4.8} & \textcolor{downc}{$-$0.8} & \textcolor{upc}{+2.0} \\
Self-check & \textcolor{upc}{+3.0} & \textcolor{downc}{$-$0.3} & \textcolor{upc}{+1.4} \\
Entailment model & \textcolor{upc}{+3.7} & \textcolor{downc}{$-$0.2} & \textcolor{upc}{+1.7} \\
\oursrow Support test & \textcolor{upc}{+5.9} & \textcolor{upc}{+1.3} & \textcolor{upc}{+3.6} \\
\bottomrule
}
\caption{Gain over sampling when CAD replaces selected answers on free-form tasks. Four ranked gates match the support test's selection count. Mean averages NQ-Swap and Ord. Color indicates 95\% intervals excluding zero.}
\label{tab:gate}
\end{table}

%% file: arxiv/sections/analysis.tex
\section{Analysis}
\label{sec:analysis}

\input{arxiv/tables/robust}
\input{arxiv/tables/examples}

\paragraph{Beyond the main setting.}
\aor{} outperforms the baselines in average score across the comparisons in Table~\ref{tab:robust}.
The lead remains when we exclude FaithEval and when the baselines receive the same memory veto, indicating that neither the multiple-choice task nor the veto alone accounts for the advantage.
Across model families, we find that \aor{} improves on sampling on OLMo-2-13B and Yi-1.5-9B, where most contrastive baselines score below sampling.
Its lead over the best baseline grows from about 3 points for Qwen3 models up to 8B to \scGainBestThirtyTwo{} points at 32B, as Figure~\ref{fig:analysis}b shows.
This suggests that checking which answers need repair remains useful as model size increases.

With the decoder settings retuned for greedy decoding, \aor{} retains its lead over the baselines.
It also improves over sampling when answers are given as one sentence, as detailed in Appendix~\ref{app:sentgrd}.
Under two LLM judges that accept paraphrased answers, \aor{} ranks highest on average, suggesting that its gains are not explained by word-overlap scoring alone.

\paragraph{Where \aor{} acts.}
\aor{} accepts \accShareORDlo{} to \accShareORDhi{}\% of initial answers on ordinary passages and \rtAccShareSwap{}\% on NQ-Swap, as Figure~\ref{fig:analysis}a shows.
It returns them unchanged, without repair decoding or a closed-book draw.
Across free-form tasks, \aor{} finds a supported repair for more than half of the rejected initial answers.
On most FaithEval questions, the passage contains no option text or multiple option texts, so no answer passes the support test and \aor{} returns a fallback answer.
In these cases, the memory veto can still mask the closed-book option during repair decoding when the passage favors another.
Table~\ref{tab:examples} illustrates acceptance, CAD and ID repairs, and the memory veto.

%% file: arxiv/tables/robust.tex
% generated by scripts/v8_numbers.py (restructure 2026-10-08); split=test
\begin{table*}[t]
\centering\small
\fulltable{\textwidth}{13}{lcccccccccccc}{%
\toprule
 & \multicolumn{2}{c}{\hd{Main tasks}} & \multicolumn{2}{c}{\hd{Held-out data}} & \multicolumn{4}{c}{\hd{Further models}} & \multicolumn{2}{c}{\hd{Other settings}} & \multicolumn{2}{c}{\hd{LLM judge}} \\
\cmidrule(lr){2-3}\cmidrule(lr){4-5}\cmidrule(lr){6-9}\cmidrule(lr){10-11}\cmidrule(lr){12-13}
\hd{Method} & \hd{No FaithEval} & \hd{With veto} & \hd{CF} & \hd{Ord.} & \hd{OLMo} & \hd{Yi} & \hd{32B} & \hd{70B} & \hd{Greedy} & \hd{Distractors} & \hd{Qwen} & \hd{Gemma} \\
\midrule
Sampling & 65.8 & 70.2 & 63.1 & \underline{83.7} & 68.6 & 67.9 & 72.3 & 68.9 & 66.6 & 63.3 & -- & -- \\
CAD & 67.5 & 71.7 & 66.6 & 81.9 & 60.1 & 64.2 & \underline{74.4} & 68.9 & 67.3 & 63.6 & $-$1.5 & +1.1 \\
AdaCAD & 68.1 & 71.6 & 65.7 & \underline{83.7} & 68.5 & 67.5 & 74.0 & 71.6 & 67.6 & 64.9 & \underline{+0.6} & +1.6 \\
COIECD & 67.5 & 71.8 & 66.7 & 82.0 & 60.0 & 64.6 & 74.3 & 69.6 & 66.5 & 63.5 & $-$1.2 & +1.3 \\
NWCAD & 66.0 & 68.6 & 63.4 & \underline{83.7} & 68.7 & 67.8 & 72.3 & 69.1 & 66.6 & 63.5 & 0.0 & +0.1 \\
RPCD & 67.7 & 72.2 & 67.8 & 83.2 & 63.5 & \underline{68.2} & 73.5 & 72.7 & 67.8 & 65.8 & $-$0.8 & +1.1 \\
ID & \underline{69.5} & \underline{72.4} & 69.4 & 83.2 & 64.3 & 56.1 & 72.8 & 66.4 & 67.6 & \underline{66.2} & +0.5 & \underline{+2.1} \\
Strict prompt & 67.5 & 70.8 & 66.7 & \underline{83.7} & \underline{70.7} & 67.3 & 74.2 & \underline{73.4} & 68.3 & 63.9 & +0.5 & +1.2 \\
Strict CAD & 67.9 & 71.3 & \underline{69.5} & 80.9 & 59.4 & 61.4 & 73.1 & 68.8 & \underline{69.1} & 62.5 & $-$2.0 & +1.1 \\
\oursrow \aor{} (ours) & \textbf{71.1} & \textbf{74.9} & \textbf{70.4} & \textbf{84.9} & \textbf{75.0} & \textbf{75.3} & \textbf{80.0} & \textbf{78.2} & \textbf{73.9} & \textbf{68.4} & \textbf{+2.1} & \textbf{+3.4} \\
\bottomrule
}
\caption{Beyond the main setting: Mean, or group average for held-out data. No FaithEval: NQ-Swap, ordinary average; With veto: veto for all; Greedy: four models, retuned, first row greedy; Distractors: passages of four other questions; Judge: free-form Mean gain over sampling.}
\label{tab:robust}
\end{table*}

%% file: arxiv/tables/examples.tex
% test answers of seed 0, results/v7/<model>/<task>/test/{base,cfg-noctx-g1.0,aor-ci-k2-r5}__s0.jsonl (uids 308, 912, 3674, 948)
\begin{table*}[t]
\centering\fontsize{7.2}{8}\selectfont
\setlength{\tabcolsep}{4pt}
\begin{tabularx}{\textwidth}{@{}>{\raggedright\arraybackslash}p{1.75cm}>{\raggedright\arraybackslash}X>{\raggedright\arraybackslash}p{3.0cm}>{\raggedright\arraybackslash}p{1.3cm}>{\raggedright\arraybackslash}p{1.2cm}>{\raggedright\arraybackslash}p{1.95cm}@{}}
\toprule
\hd{Task, model} & \hd{Passage, excerpt} & \hd{Question} & \hd{Sampling} & \hd{CAD} & \hd{\aor{}} \\
\midrule
NQ-Orig.\newline Qwen3-8B & UConn owns the two longest winning streaks \dots The longest streak, 111 straight wins, started \dots & how many games in a row have the uconn women's basketball team won & 111 \cmark & two \xmark & 111 \cmark\newline accepted \\
\addlinespace
NQ-Swap\newline Qwen3-8B & In 1971, after the Irish War of Independence \dots most of Ireland seceded from the United Kingdom \dots & when did ireland gain its independence from england & 1922 \xmark & 1971 \cmark & 1971 \cmark\newline CAD repair \\
\addlinespace
NQ-Swap\newline Llama-3.1-8B & Neal Dahlen as Hiccup Horrendous Haddock III, the son of the Viking chief Stoick the Vast and Valka. & who voices hiccup in how to train your dragon 2 & Jay Baruchel \xmark & Jay Baruchel \xmark & Neal Dahlen \cmark\newline ID repair \\
\addlinespace
FaithEval\newline Qwen3-14B & \dots recent studies have shown that minerals hold a pivotal role in the fossilization process. \dots & Which of the following would most likely form a fossil? A.~a rock, B.~a crystal, C.~a mineral, D.~a bone & D \xmark & D \xmark & C \cmark\newline D vetoed \\
\bottomrule
\end{tabularx}
\caption{Test answers from three models, with passage excerpts, illustrating acceptance, CAD and ID repairs, and the memory veto. A check marks a correct answer; labels below \aor{}'s answers indicate the action taken.}
\label{tab:examples}
\end{table*}

%% file: arxiv/sections/conclusion.tex
\section{Conclusion}
\label{sec:conclusion}

We propose Accept-or-Repair, a training-free procedure that keeps passage-supported initial answers and uses contrastive decoding to repair unsupported ones.
AoR also applies the memory veto to reject repairs that repeat the answer generated without the passage when the passage favors another answer.
AoR improves average scores over plain sampling on counterfactual and ordinary passages across ten models.
In the main evaluation on six models and five benchmarks, it yields large gains on counterfactual passages and small gains on ordinary ones.
Replacing only unsupported initial answers improves the average score of every decoder in this evaluation, compared with using that decoder to answer every question.
Our findings suggest that checking passage support helps models follow passages that contradict their knowledge while preserving accuracy on passages that agree with it.

%% file: arxiv/sections/closing.tex
\section*{Limitations}

\aor{} uses word matching to check passage support for short answers and single sentences.
Long-form or abstractive answers call for support tests that account for paraphrasing and information combined across the passage.
We examine model self-checking and an entailment model as alternative tests on short-answer tasks in Appendix~\ref{app:further}.
Our experiments cover English benchmarks and models of up to 70B parameters.

The contrastive repair decoders require access to the model's next-token distributions.
We also evaluate strict-prompt resampling as an alternative repair generator in Section~\ref{sec:results}.

\section*{Ethics Statement}

All datasets and models are public, and we use them for research under the terms of their releases: NQ-Swap and ConFiQA under the MIT license, SQuAD and HotpotQA under CC BY-SA 4.0, SciQ under CC BY-NC 3.0, Qwen3, Qwen2.5-32B-Instruct, Gemma 4, Mistral-7B-Instruct, OLMo-2 and Yi-1.5 under Apache-2.0, Llama-3.1 under the Llama 3.1 Community License, and the DeBERTa-v3 entailment model under the MIT license.

\aor{} helps models base their answers on the provided passage.
Applications should assess source reliability, since passage support alone does not establish factual correctness \citep{lee2022norose}.

%% file: arxiv/sections/appendix.tex
\definecolor{groupbg}{HTML}{FFF2CC}
\section{Prompts and Decoders}
\label{app:prompts}

Initial answers and contrastive decoders use single-turn user prompts rendered with the model's chat template.
We add no system message, and the Llama-3.1 template adds its default system header.
The Qwen3 models run with thinking disabled.

Table~\ref{tab:prompts} gives the prompts of the free-form tasks.
The instruction, passage and question are separated by blank lines, with the labels Passage and Question before the corresponding text.
For HotpotQA, the label is Passages, directives refer to multiple passages, and each paragraph begins with its title and a colon.
For the multiple-choice tasks the directive reads ``Answer the multiple-choice question according to the passage'', the format sentence asks for only the letter of the correct option, and the question is followed by the options, one on each line in the form ``A. text''.
The contrastive prompts make the same changes to the instruction or passage as their free-form counterparts in Table~\ref{tab:prompts}.
The closed-book prompt reads ``Answer the multiple-choice question. Reply with only the letter of the correct option.'' followed by the question and the options.

CAD, Strict CAD, RPCD and ID score a token $y$ at each step by $(1+\gamma)\log p(y\mid x)-\gamma\log p(y\mid x')$, where $x$ is the prompt with the passage, $x'$ is the contrasted prompt in Table~\ref{tab:prompts} and $\gamma$ is the contrast strength.
They sample from this score using the model's temperature and truncation settings.
CAD and Strict CAD contrast with the closed-book prompt, RPCD with the passage of another example and ID with the opposite directive.
Strict CAD uses the strict prompt as $x$.
For RPCD, we take the passage from the example $\lfloor n/3\rfloor+1$ places later in the same pool of $n$ examples, wrapping around at the end.

AdaCAD scales $\gamma$ at every step by the Jensen--Shannon divergence between the two distributions in bits.
COIECD adds the contrast to the log probability under $x$ when a token violates its entropy constraint, and to the closed-book log probability otherwise.
NWCAD uses the closed-book distribution when it is confident and close to the distribution under $x$.
Otherwise, it uses the distribution under $x$ when that distribution is confident, and the CAD score when neither condition holds.
Confidence is measured by the gap between the two highest token probabilities.
We use the paper's margins of 0.30 for the closed-book distribution and 0.05 for the distribution under $x$.
We implement COIECD and NWCAD from their papers and also consult the released COIECD code.

\begin{table*}[!t]
\centering\footnotesize
\renewcommand{\arraystretch}{1.25}
\setlength{\aboverulesep}{0pt}\setlength{\belowrulesep}{0pt}
\begin{tabularx}{\textwidth}{>{\raggedright\arraybackslash}p{0.13\textwidth}>{\raggedright\arraybackslash}X>{\raggedright\arraybackslash}p{0.25\textwidth}}
\toprule
\hd{Prompt} & \hd{Text} & \hd{Used by} \\
\midrule
\rowcolor{oursbg}Standard & \texttt{Answer the question using the passage. Reply with only the answer, a few words at most.} + passage + question & sampling, \aor{}, every contrastive decoder except Strict CAD \\
\rowcolor{oursbg}Closed-book & \texttt{Answer the question. Reply with only the answer, a few words at most.} + question & the memory veto of \aor{}, contrast of CAD, AdaCAD, COIECD, NWCAD and Strict CAD \\
Other passage & standard instruction + passage of another example + question & contrast of RPCD \\
Opposite directive & \texttt{Answer the question using your own knowledge and ignore the passage. Reply with only the answer, a few words at most.} + passage + question & contrast of ID \\
Strict & \texttt{Answer the question using only the passage, even if the passage contradicts what you know. Reply with only the answer, a few words at most.} + passage + question & Strict prompt, Strict CAD \\
Copy & \texttt{Answer the question using the passage. Reply with only the answer, copied from the passage in as few words as possible.} + passage + question & copy prompt, alone and with CAD \\
Feedback & standard prompt + plain answer + \texttt{Your answer does not occur in the passage. Answer the question again using the passage. Reply with only the answer, a few words at most.} & feedback repair \\
\bottomrule
\end{tabularx}
\caption{Prompts of the free-form tasks. A plus sign joins the parts of a prompt. Shading marks the two prompts that \aor{} samples from. Its repairs use the prompts of CAD and ID.}
\label{tab:prompts}
\end{table*}

\section{Support Test and Veto}
\label{app:support}

\paragraph{Support test.}
For short free-form answers, the support test reads the first line.
Normalization lowercases the answer and passage, replaces ASCII punctuation with spaces and preserves other punctuation and symbols as separate tokens.
It then removes the articles a, an and the, and collapses whitespace.
A normalized answer passes when it is nonempty and occurs in the normalized passage as a contiguous sequence of whole tokens.
Answers that normalize to yes or no also pass.
For multiple-choice questions, we extract the option letter as in evaluation and apply the same normalized token matching to each option's text.
The answer passes only when the selected option is the sole match.

\paragraph{Memory veto.}
We sample one closed-book answer $m$ for each question and reuse it across the three runs.
For free-form questions, the veto rejects repairs that match $m$ after normalization if $m$ fails the support test.

For multiple-choice questions, an option scores 2 when its text passes the token-matching rule above.
Otherwise, its score is the fraction of its normalized words, excluding stop words, that occur in the normalized passage.
If another option scores higher than the option selected by $m$, we mask tokens representing the letter of $m$ by setting their scores to $-\infty$ at every repair decoding step.
These tokens are identified by stripping spaces, tabs, line breaks, parentheses, periods and colons from the ends of their decoded text and checking whether the remainder equals the letter.
If no option can pass both the support test and the veto, \aor{} returns the first repair that names a non-vetoed option as its fallback and stops generating repairs.

\paragraph{Cost and latency.}
Repairs and the closed-book answer are generated only for unsupported initial answers.
The token cost in Table~\ref{tab:main} counts prompt and generated tokens across decoding calls, including tokens processed under both prompts in contrastive decoding.

We measure latency with Qwen3-8B at batch size one on an A100 GPU, using 100 test examples from each ordinary task.
For these free-form questions, we generate the closed-book answer alongside the CAD repair because generating the repair does not depend on the closed-book answer.
\aor{} takes \latOursORD{} times as long as sampling, compared with \latCadORD{} for CAD.

\section{Evaluation Details}
\label{app:eval}

\paragraph{Data.}
Table~\ref{tab:data} reports the sizes of the data splits.
For the main tasks, we use the validation splits of SQuAD and HotpotQA's distractor setting, the development split of NQ-Swap, which also supplies NQ-Original, and the counterfactual test split of FaithEval version 1.0.
The held-out tasks use the test splits of the three ConFiQA subsets, the TriviaQA web development set from MRQA 2019, and SciQ's validation and test splits with their support paragraphs.
For ConFiQA, we keep questions whose changed answer shares no alias with the original answer and whose passage has at most 1,000 words.
We score F1 against the answer stated in the passage.
The original versions pair the same questions with the unchanged passages and answers.
We randomly sample 1,000 examples from each task, reserving 200 for development and 800 for testing.
We use development examples to select decoding settings and AoR's repair sequence, keeping model weights fixed.

\paragraph{Scoring.}
We score the first line of short free-form answers with token-level F1 after lowercasing and removing punctuation and articles, taking the highest score across reference answers.
For multiple-choice tasks, we score accuracy after extracting an option letter.
We check for a valid option letter first at the start of the response, then after ``answer'', ``option'' or ``choice'', and finally in parentheses or an option label at the start of a line.
Responses without a recognized option letter are incorrect.
Generation stops after at most 48 new tokens for short free-form answers and 32 for multiple-choice answers.

\paragraph{Sampling.}
All runs use bfloat16 weights, PyTorch \citep{paszke2019pytorch} and Transformers \citep{wolf-etal-2020-transformers} on RTX A6000, A100 and H100 GPUs.
Table~\ref{tab:sampling} lists the sampling settings for each model.
To pair sampled outputs across methods, we generate Gumbel noise with a 32-bit integer hash of the run seed, example identifier, decoding step and token index.
After contrastive adjustment, temperature scaling and filtering, we add this noise to the token scores and select the token with the highest score.
This samples from each method's distribution while giving methods with identical distributions the same token.
\aor{} uses the initial answer's noise for its repairs, so its CAD repair matches the CAD baseline unless the memory veto masks an option.
A second draw from the same decoder uses the next run's noise.

\paragraph{Significance.}
For the main comparison, confidence intervals use 2,000 bootstrap resamples of questions, except for SQuAD, where we resample passages together with all their questions.
The resampling indices are shared across models and runs, and across the paired NQ-Swap and NQ-Original tasks.
For each resample, we average score differences over the tasks in each group and take the mean of the two group differences to compute Mean.
Task-level comparisons use a Holm correction.

\input{arxiv/tables/data}
\input{arxiv/tables/sampling}

\section{Baseline Settings}
\label{app:strengths}

Table~\ref{tab:strengths} reports development gains over sampling for each setting, averaged across the counterfactual and ordinary groups.
We select the setting with the highest average gain, retaining the original paper's recommendation when it is within 0.1 points of the best.
Strict CAD uses CAD's selected strength.

Table~\ref{tab:gridtest} reports test gains across these settings, with Strict CAD evaluated at the same strengths as CAD.
Gains on ordinary passages remain small, and the settings with the largest counterfactual gains lower ordinary scores.
This pattern suggests that the trade-off is not confined to the selected settings.

For NWCAD, we trace decoding on 200 test examples from each task using Qwen3-8B at the selected setting.
The share of steps using the CAD score reaches a maximum of \nwcadCadMax{}\% across tasks, with the remaining steps sampling from the prompt with or without the passage.

\input{arxiv/tables/strengths}
\input{arxiv/tables/gridtest}

\section{Further Analyses}
\label{app:further}

\paragraph{Main comparison.}
Against Strict CAD, the strongest baseline, \aor{} gains \oursVsStrictCadCF{} points on counterfactual and \oursVsStrictCadORD{} on ordinary passages.
Every baseline trails it significantly in both groups, and its smallest lead in Mean on one of the six models is \minGainModel{} points.
On individual tasks, \taskWins{} of the \taskCells{} gains of \aor{} over a baseline remain significant after Holm correction.
Without FaithEval, it leads the closest baseline, \ffMinName{}, by \ffMinLead{} points in Mean.

\paragraph{Supported answers.}
In Table~\ref{tab:anatomy}, support is determined by the test of \aor{}.
An answer is correct when it matches a gold answer exactly, or for FaithEval when it names the right option.
On ordinary passages, \dcPSORD{}\% of the plain answers are supported and score \dcPlainSORD{} F1.
On these answers CAD scores \dcCondCadORDMag{} points lower and Strict CAD \dcCondStrictCadORDMag{}.
CAD changes \chgCadPct{}\% of them, and \chgCadUnsup{}\% of these changes give an answer that the passage does not contain.
The \oursBreakCountORD{} correct plain answers that \aor{} breaks on ordinary passages are unsupported because the passage contains the gold answer only inside a longer word, as \emph{Ghana} inside \emph{Ghanaian}.

\paragraph{Precision of the support test.}
Of the plain answers that the test accepts on ordinary passages, \accPrecORD{}\% have an F1 of at least 0.5.
Using this threshold to define correctness, every baseline breaks more supported initial answers than it corrects, from \accRatioLo{} times as many for \accRatioLoName{} to \accRatioHi{} times for \accRatioHiName{}.
Within this supported subset, retaining the initial answers improves the binary score over the decoder exactly when this ratio exceeds one.
Across free-form tasks, \precSupLo{} to \precSupHi{}\% of supported plain, CAD and ID draws have an F1 of at least 0.5, compared with \precUnsLo{} to \precUnsHi{}\% of unsupported draws.
\accMemSwap{}\% of the plain answers that \aor{} accepts on NQ-Swap equal the closed-book answer.
On the \mPsgLo{} to \mPsgHi{}\% of ordinary examples whose passage also contains a wrong closed-book answer, \aor{} scores highest on each ordinary task.

\paragraph{Repair term.}
For \aor{} itself, $\Delta_{\neg S}$ of Equation~\ref{eq:split} splits into a repair term and a fallback term,
\begin{equation}
\Delta_{\neg S}=\rho\,(\pi-u_R)+(1-\rho)(\phi-u_F),
\end{equation}
where $\rho$ is the share of unsupported answers that a supported repair replaces, $\pi$ and $\phi$ are the mean scores of the supported repairs and of the fallbacks, and $u_R$ and $u_F$ those of the plain answers they replace.
\aor{} thus gains over sampling exactly when the repair term $\rho\,(\pi-u_R)$ exceeds the fallback loss $(1-\rho)(u_F-\phi)$.
On the free-form tasks, the supported repairs of \aor{} score \repPiLo{} to \repPiHi{} F1, where the plain answers they replace score \repURLo{} to \repURHi{}.
The repair term is \repTermLo{} to \repTermHi{} points, while the fallback loss stays within \fbTermMaxMag{} points of zero.
On ordinary passages, \aor{} raises the F1 of \oursRaiseLow{} answers for each one it lowers, and no baseline raises more than \baseRaiseLowHi{}.

\paragraph{Where contrastive decoders break answers.}
Table~\ref{tab:mech} gives the share and the plain F1 of each bin of Figure~\ref{fig:mech}, with the probability taken under the closed-book prompt.
Strict CAD, CAD and COIECD lose more F1 in the higher closed-book probability bins, while AdaCAD, which weakens the contrast where the two predictions agree, loses little across bins.
CAD, COIECD and Strict CAD break correct supported answers more often when the closed-book answer is also correct.
With correctness defined as in Table~\ref{tab:anatomy} and the six models pooled, CAD breaks \memAgreeCad{}\% of these answers, compared with \memDifferCad{}\% when the closed-book answer is wrong.
The corresponding rates are \memAgreeCoiecd{} and \memDifferCoiecd{}\% for COIECD, and \memAgreeStrict{} and \memDifferStrict{}\% for Strict CAD.
RPCD and ID, whose contrasting prompt also contains a passage, show the opposite pattern, with rates of \memAgreeRpcd{} versus \memDifferRpcd{}\% for RPCD and \memAgreeId{} versus \memDifferId{}\% for ID.
On ordinary passages, a second plain sample breaks \resampleBrkORD{}\% of correct initial answers, compared with \cadBreakORD{}\% for CAD using the same noise as the initial sample.

\paragraph{Wrapping other decoders.}
Every gain in Mean of Table~\ref{tab:plugin}, from \plugMeanLo{} to \plugMeanHi{} points, is significant.
Draws that state no answer are rare, so $\epsilon$ in Equation~\ref{eq:wrap} stays below \wrapResidMax{} points.
The counterfactual score of each wrapped decoder stays within \plugCFAbsMax{} points of its score when used alone.

\paragraph{Ablations.}
Table~\ref{tab:abltask} gives six variants of Table~\ref{tab:ablation} on each task, together with a fallback to the plain answer.
Without the initial acceptance test, \aor{} still scores \noAcceptORDAbs{} on ordinary passages, above sampling.
The support checks on repairs remain in place.
Accepting an answer that some span of the passage matches with a token F1 of at least 0.5 or 0.8 gives a Mean of \grHalfMean{} or \grEightMean{}.
On \grMemHalf{}\% of NQ-Swap questions, the 0.5 threshold accepts an initial answer containing the answer from the original passage.
CAD or ID alone lowers the Mean by \ablKOneCMeanMag{} points, and at the 16 combinations of CAD and ID strengths the Mean lies between \gridMinMean{} and \gridMaxMean{}.

\paragraph{Memory veto.}
Removing the memory veto lowers FaithEval accuracy by \ablNoVetoFaithMag{} points and the counterfactual average by \ablNoVetoCFMag{} points.
Table~\ref{tab:faith} separates the veto from the decoders on FaithEval, and Table~\ref{tab:vetobase} gives the veto to every baseline.
The first row of Table~\ref{tab:faith} picks the option with the highest veto score and splits ties evenly.
A decoder with the veto returns the answer of its run with the masked letter where the veto holds and its usual answer elsewhere.
On free-form tasks, the veto does not alter a baseline's answer because it filters only the fallback candidates used by \aor{}.
The baselines with the veto trail \aor{} by at least \oursVsVetoMinCF{} points on counterfactual and \oursVsVetoMinORD{} on ordinary passages.

\paragraph{Repairs without contrast.}
With repairs sampled from the strict prompt, \aor{} achieves a Mean of \strictResampleMeanAbs{}, \strictResMinLead{} points above Strict CAD.
On NQ-Swap, the gain over sampling is \strictResSwapGain{} points, compared with \oursSwapGain{} with contrastive repairs.

\paragraph{Majority voting.}
The voting variants in Table~\ref{tab:ablation} select the most frequent answer after applying the support test's normalization.
One votes among the plain answers of three runs, and the other among the plain, CAD and ID answers of one run.
Both return the initial plain answer when no answer occurs twice.

\paragraph{Feedback repair.}
The feedback repair adds the unsupported plain answer as an assistant turn and the feedback instruction from Table~\ref{tab:prompts} as a new user turn, then samples another answer with the same noise.
Asked again, the model repeats \feedSame{}\% of the rejected answers, after the normalization of the support test, and \feedSup{}\% of its new answers are supported, against \feedCadSup{}\% of the CAD answers to the same questions.
As the only repair of \aor{}, feedback gains \feedGainSwap{} points over sampling on NQ-Swap and \feedGainORD{} on ordinary passages.
A CAD repair gives larger gains of \oneCadGainSwap{} and \oneCadGainORD{} points, respectively, at a similar token cost.

\paragraph{Spans of the passage.}
Table~\ref{tab:span} compares other ways to answer in the words of the passage, and \aor{} with one repair, on the four free-form tasks, since a multiple-choice answer is a letter.
Span decoding samples the plain answer under the constraint that it is a span of the passage.
At each step, it considers the 64 highest-probability tokens and keeps those that extend a prefix of a passage span starting at a word boundary, or a prefix of yes or no.
The answer can end only at a complete passage span or at yes or no.
If no token satisfies the constraint before the answer is complete, decoding continues without the constraint.
The copy prompt of Table~\ref{tab:prompts} asks for the answer copied from the passage, for sampling and for CAD, which keeps its closed-book prompt.
The copy prompt lowers the ordinary score by \copyLossORDMag{} points, and span decoding trails \aor{} by \oursVsSpanSwap{} points on NQ-Swap and \oursVsSpanORD{} on ordinary passages.

\paragraph{Span-restricted repair.}
Span Strict CAD in Table~\ref{tab:ablation} is Strict CAD decoded under the constraint of span decoding.
Among six repair sequences with a span-restricted draw that we compared on the development split, the sequence that ends with it scores highest on NQ-Swap and on the ordinary tasks.
On the test split it raises NQ-Swap from \spRepSwapOurs{} to \spRepSwap{} and the Mean of each of the six models.
Without the constraint, Strict CAD as the third repair raises the Mean by \ablCicpMean{} points.

\paragraph{Gates.}
Table~\ref{tab:gate} compares gates for selecting which initial answers to replace on the four free-form tasks.
All gates use the CAD answer from the same run as the only repair and return it without a support check.
For each model, task and run, four gates rank initial answers and flag the same number as the support test rejects.
The first two rank by mean token log-probability and by the gain in log-probability from adding the passage to the prompt.
The third uses the probability of yes when the answering model is asked whether the passage supports its answer \citep{kadavath2022language}.
The fourth uses entailment probability from DeBERTa-v3-large \citep{he2023debertav3} fine-tuned for natural language inference by \citet{laurer2023less}.
Each gate flags the answers with the lowest scores.

The self-check prompt reads ``Read the passage and decide whether it supports the proposed answer to the question.'', followed by the passage, the question and the answer and by ``Does the passage support the proposed answer? Reply with only yes or no.'', in the plural on HotpotQA.
The probability of yes is taken against that of no at the first token of the reply.
The entailment model reads overlapping 400-token windows of the passage as premises and takes the highest entailment probability across windows.
The hypothesis is the answer, preceded by the question when the answer has at most three words.
The closed-book gate flags the plain answers that equal the closed-book answer, \gateMemRate{}\% of the answers against \gateRate{}\% for the support test.

The Mean of Table~\ref{tab:gate} averages NQ-Swap and the ordinary average.
The support test raises this Mean by \gateSupMean{} points, significantly more than the other gates, which gain at most \gateOtherMeanHi{} points.
Under the two LLM judges, its gains over sampling are \gateJudgeSupQ{} and \gateJudgeSupG{} points, compared with at most \gateJudgeOtherQ{} and \gateJudgeOtherG{} for the other gates.
Its lead over each gate is at least \gateJudgeMinMean{} points, with 95\% intervals above zero.
With the self-check or the entailment model as its test, \aor{} stays \semSelfMinLead{} and \semNliMinLead{} points above the closest baseline in Mean.

\paragraph{Held-out data and further models.}
On the counterfactual subsets of ConFiQA, \aor{} gains \hoGainBase{} points over sampling, significantly more than every baseline.
Across the held-out ordinary tasks, \aor{} gains \agGainBase{} points over sampling on average, while no baseline improves significantly.
On OLMo-2-13B, Yi-1.5-9B, Qwen3-32B and Llama-3.1-70B, \aor{} gains significantly over sampling in both groups.
Its Mean exceeds the closest baseline by \olmoGainBest{}, \yiGainBest{}, \bigGainBest{} and \hugeGainBest{} points, respectively.
Tuning every baseline on the development split of each of OLMo-2-13B, Yi-1.5-9B and Qwen3-32B leaves the closest one \rtOlmoGap{}, \rtYiGap{} and \rtBigGap{} points behind \aor{}.
On the Qwen3 models of Figure~\ref{fig:analysis}b, the supported answers cost the best baseline \scLossBestThirtyTwo{} points of its Mean at 32B, against \scLossBestEight{} at 8B.

\paragraph{Passage length and distractors.}
Split into quartiles by the length of the passage, \aor{} gains \lenGainLo{} to \lenGainHi{} points over sampling in every quartile of every free-form task, including the longest quartile of HotpotQA, whose passages have at least \lenHotTop{} words.

For Table~\ref{tab:fivepsg}, every question of NQ-Swap and NQ-Original keeps its passage and receives the passages of four other NQ-Swap questions, in the same version, that contain neither its gold answer nor its original answer as whole words.
The question's own passage occupies a random one of the five positions, shared across its original and counterfactual versions along with the distractors.
We separate passages by blank lines and use the HotpotQA prompts.
This setting runs once, with the strengths of the main runs.
\aor{} scores highest on both tasks, indicating that its lead persists with the added distractors.

\paragraph{Other scorers.}
Table~\ref{tab:contain} scores a free-form answer as correct when a gold answer occurs in it as whole words after the normalization of the scorer.
For Table~\ref{tab:judge}, Qwen2.5-32B-Instruct \citep{yang2024qwen2}, a model outside the main comparison, judges the free-form answers with greedy decoding.
It reads ``You grade answers to a reading-comprehension question. A candidate answer is correct if it means the same as one of the reference answers, even when it is worded differently, for example with an alias, an abbreviation, a number written in words, or extra words around the answer. It is wrong if it names something else or adds a claim that contradicts the references. Judge only against the references.'', followed by the question, the gold answers and the answer, and replies with yes or no.
When a method's answer equals the plain answer of the same run after the support test's normalization, it receives the same judgment.
Differences from sampling therefore reflect changed answers alone.

Among 300 random plain answers with an exact match, the judge accepts \judgeCheckEM{}\%, and among 300 with an F1 of zero, \judgeCheckFzero{}\%.
The second judge, Gemma-4-31B-it \citep{gemmateam2026gemma4}, reads the same prompt.
A reply counts as yes when its first word is yes.
It accepts \jgCheckEM{}\% and \jgCheckFzero{}\% of the same two samples.

The judge columns of Table~\ref{tab:robust} report gains over sampling in the Mean of NQ-Swap and the ordinary average.
Intervals for judged differences use 2,000 resamples of the questions in each task.

Of the \joN{} judged NQ-Swap answers that equal the question's original answer, Qwen2.5-32B-Instruct accepts \joQwen{} and Gemma-4-31B-it \joGemma{}.
Under Qwen2.5-32B-Instruct, \aor{} gains \judgeSwap{} points over sampling on NQ-Swap and \judgeORD{} on ordinary passages, both significantly, and it leads every baseline significantly on both.
Under Gemma-4-31B-it, \aor{} gains \jgSwap{} points on NQ-Swap, more than any other method, while CAD loses \jgCadORDMag{} and Strict CAD \jgStrictCadORDMag{} points on ordinary passages.
Under this judge, the wrapper of Table~\ref{tab:plugin} raises the judged ordinary accuracy of the seven decoders that lose on ordinary passages by \wrapJudgeOrdLo{} to \wrapJudgeOrdHi{} points.

Under the two judges, \aor{} breaks \jbBrkCorOursORD{}\% and \jgBrkCorOursORD{}\% of the ordinary plain answers that the judge accepts, and CAD \jbBrkCorCadORD{}\% and \jgBrkCorCadORD{}\%.
Of the ordinary plain answers that the test rejects but a judge accepts, \aor{} replaces \rejOkQ{}\% and \rejOkG{}\% by answers that the same judge accepts.

Qwen2.5-32B-Instruct rejects \joGoldQRej{}\% of the NQ-Swap answers that equal the substituted gold answer, and Gemma-4-31B-it \joGoldGRej{}\%.
With such exact answers counted as correct, \aor{} gains \emSwapQ{} points on NQ-Swap under the first judge.

\input{arxiv/tables/mech}
\input{arxiv/tables/abltask}
\input{arxiv/tables/faithdecomp}
\input{arxiv/tables/vetobase}
\input{arxiv/tables/span}
\input{arxiv/tables/fivepsg}
\input{arxiv/tables/contain}
\input{arxiv/tables/judge}

\section{Sentence Answers and Greedy Decoding}
\label{app:sentgrd}

\paragraph{Sentence answers.}
We evaluate sentence answers on NQ-Swap, NQ-Original and SQuAD using the same examples and gold answers as the short-answer experiments.
We replace the format instruction in Table~\ref{tab:prompts} with ``Reply with one short sentence.'' and allow up to 64 new tokens.
An answer is correct when a gold answer occurs in its first line as whole words after scorer normalization.

The support test normalizes the answer's first line and the passage as in Appendix~\ref{app:support}, then removes question words and function words from the answer.
The answer passes when at least one word remains and all remaining words occur in the passage.
Answers normalized to yes or no also pass.
We selected the requirement to match all remaining words using the development split.
We generate the closed-book answer in the same sentence format and retain the short-answer procedures for the memory veto, repairs and fallback.

We find that \aor{} improves over sampling on all three tasks in Table~\ref{tab:sent}, with a larger gain on NQ-Swap than on the ordinary tasks.

\paragraph{Greedy decoding.}
We run each method once on each task for the four models in Table~\ref{tab:greedy}, choosing the token with the highest score at each step.
We reselect decoder settings by the rule in Section~\ref{sec:setup}, evaluating the grid in Table~\ref{tab:strengths} with greedy decoding on the four models' development split.
The chosen strengths are 0.5 for CAD, Strict CAD and RPCD, and 1 for AdaCAD and ID.
NWCAD uses a divergence threshold of 0.1, and COIECD uses strength 1 and a constraint scale of 0.1.
\aor{} generates its initial answer and CAD/ID repairs greedily at these settings and reuses the sampled closed-book answer from the main runs to keep the memory veto's reference answer fixed.

Under greedy decoding, \aor{} gains \grdGainBaseMean{} points in Mean over plain greedy decoding and outperforms the baselines on each model in Table~\ref{tab:greedy}.
Together, these results suggest that \aor{}'s gains extend beyond short answers generated by sampling.

\input{arxiv/tables/sent}
\input{arxiv/tables/greedy}

\section{Examples}
\label{app:examples}

Table~\ref{tab:examples} illustrates how \aor{} accepts an initial answer, returns a CAD or ID repair, and applies the memory veto.
On the ordinary passage, CAD answers \emph{two}, a count of winning streaks, to a question about consecutive wins.
\aor{} keeps the supported initial answer 111.

On the first NQ-Swap passage, \aor{} replaces the initial answer 1922 with CAD's answer 1971, the year given in the passage.
On the second, CAD repeats Jay Baruchel, whose name is absent from the passage.
\aor{} rejects this repair and returns the ID answer Neal Dahlen, which occurs in the passage.

On FaithEval, sampling and CAD choose D, which is also the closed-book answer.
\aor{}'s memory veto masks D because the passage favors C, and CAD then generates C as the repair.

\section{Results by Model}
\label{app:models}

Table~\ref{tab:permodel} lists average scores for each method and each of the six models in the main comparison.

\input{arxiv/tables/permodel}

%% file: arxiv/tables/data.tex
\begin{table}[!t]
\centering\small
\fulltable{\columnwidth}{5}{llccc}{%
\toprule
\hd{Task} & \hd{Metric} & \hd{Dev.} & \hd{Test} & \hd{Used} \\
\midrule
\grouprow{5}{Counterfactual passages} \\
NQ-Swap & F1 & 200 & 800 & 724 \\
FaithEval & accuracy & 200 & 800 & 800 \\
\grouprow{5}{Ordinary passages} \\
NQ-Original & F1 & 200 & 800 & 724 \\
SQuAD & F1 & 200 & 800 & 724 \\
HotpotQA & F1 & 200 & 800 & 800 \\
\grouprow{5}{Held-out counterfactual passages} \\
ConFiQA-QA, -MR, -MC & F1 & 200 & 800 & 800 \\
\grouprow{5}{Held-out ordinary passages} \\
ConFiQA original & F1 & 200 & 800 & 800 \\
TriviaQA web & F1 & 200 & 800 & 800 \\
SciQ & accuracy & 200 & 800 & 800 \\
\bottomrule
}
\caption{Counts of development and test examples. Used excludes test examples whose question, or for SQuAD whose passage, occurs in the development split. ConFiQA rows represent each subset.}
\label{tab:data}
\end{table}

%% file: arxiv/tables/sampling.tex
\begin{table}[!t]
\centering\small
\fulltable{\columnwidth}{4}{lccc}{%
\toprule
\hd{Model} & \hd{Temperature} & \hd{Top-$p$} & \hd{Top-$k$} \\
\midrule
Qwen3 (all sizes) & 0.7 & 0.8 & 20 \\
Llama-3.1 (8B, 70B) & 0.6 & 0.9 & off \\
Mistral-7B & 0.7 & 0.9 & off \\
OLMo-2-13B & 0.7 & 0.9 & off \\
Yi-1.5-9B & 0.7 & 0.9 & off \\
\bottomrule
}
\caption{Sampling settings of every model, shared by all methods.}
\label{tab:sampling}
\end{table}

%% file: arxiv/tables/strengths.tex
% split=dev
\begin{table*}[!t]
\centering\small
\fulltable{\textwidth}{4}{llll}{%
\toprule
\hd{Method} & \hd{Parameter} & \hd{Settings tried} & \hd{Scores} \\
\midrule
CAD & strength & 0.25, 0.5, 1, 2 & 1.81, 2.67, \textbf{2.66}, $-$0.49 \\
AdaCAD & strength & 0.5, 1, 2, 4 & 1.26, \textbf{1.67}, 1.54, 0.35 \\
RPCD & strength & 0.25, 0.5, 1, 2 & 1.85, 2.75, \textbf{3.11}, 1.00 \\
ID & strength & 0.25, 0.5, 1, 2, 4, 8 & 0.92, 1.73, 2.82, \textbf{3.10}, 0.76, $-$10.89 \\
NWCAD & divergence threshold & 0.05, 0.1, 0.2, 0.3, 0.5 & 0.17, \textbf{0.23}, 0.20, $-$0.16, $-$0.76 \\
COIECD & strength, constraint scale & (0.5, 0.25), (1, 0.25), (2, 0.25), (1, 0.1), (1, 0.5) & $-$12.55, 1.19, \textbf{2.60}, 1.54, 0.51 \\
\bottomrule
}
\caption{Settings tried on the development split and their scores in the same order. Scores average the counterfactual and ordinary gains over sampling, in points. Bold marks the chosen setting.}
\label{tab:strengths}
\end{table*}

%% file: arxiv/tables/gridtest.tex
% generated by scripts/v8_numbers.py from frontier_test.json; split=test
\begin{table*}[!t]
\centering\small
\fulltable{\textwidth}{4}{llll}{%
\toprule
\hd{Method} & \hd{Settings} & \hd{Counterfactual} & \hd{Ordinary} \\
\midrule
CAD & 0.25, 0.5, 1, 2 & 3.7, 5.9, \textbf{7.8}, 8.0 & 0.2, $-$0.1, \textbf{$-$2.1}, $-$7.7 \\
AdaCAD & 0.5, 1, 2, 4 & 2.3, \textbf{3.2}, 3.7, 3.2 & 0.2, \textbf{0.1}, $-$0.5, $-$2.5 \\
RPCD & 0.25, 0.5, 1, 2 & 2.8, 4.5, \textbf{5.8}, 5.7 & 0.8, 0.7, \textbf{$-$0.3}, $-$4.2 \\
ID & 0.25, 0.5, 1, 2, 4, 8 & 1.5, 2.7, 4.5, \textbf{5.8}, 4.2, $-$7.7 & 0.4, 0.5, 0.8, \textbf{0.0}, $-$4.3, $-$15.3 \\
NWCAD & 0.05, 0.1, 0.2, 0.3, 0.5 & 0.2, \textbf{0.3}, 0.1, $-$0.1, $-$0.7 & 0.0, \textbf{0.1}, $-$0.3, $-$0.6, $-$1.2 \\
COIECD & (0.5, 0.25), (1, 0.25), (2, 0.25), (1, 0.1), (1, 0.5) & $-$22.8, 1.9, \textbf{7.7}, 2.7, 1.0 & $-$4.3, 0.2, \textbf{$-$1.9}, 0.1, 0.1 \\
Strict CAD & 0.25, 0.5, 1, 2 & 6.0, 7.9, \textbf{9.6}, 8.8 & $-$0.2, $-$1.1, \textbf{$-$3.6}, $-$10.3 \\
Strict prompt & none & \textbf{2.2} & \textbf{0.0} \\
\bottomrule
}
\caption{Gain over sampling at each setting of Table~\ref{tab:strengths}, on the counterfactual and the ordinary average of the test split. Bold marks the chosen setting.}
\label{tab:gridtest}
\end{table*}

%% file: arxiv/tables/mech.tex
% generated by scripts/v8_numbers.py from out/review_c11/mech.json; do not edit by hand
\begin{table}[t]
\centering\small
\fulltable{\columnwidth}{5}{lcccc}{%
\toprule
 & \multicolumn{4}{c}{\hd{Closed-book probability $p(a\mid q)$}} \\
\cmidrule(lr){2-5}
 & $<0.01$ & $0.01$--$0.1$ & $0.1$--$0.5$ & $\ge 0.5$ \\
\midrule
Share (\%) & 72.7 & 8.8 & 7.2 & 11.3 \\
Plain F1 & 80.5 & 85.1 & 86.3 & 89.3 \\
\grouprow{5}{F1 change} \\
CAD & $-$3.0 & $-$5.4 & $-$7.1 & $-$7.4 \\
AdaCAD & $-$1.0 & $-$1.3 & $-$1.2 & $-$0.7 \\
COIECD & $-$2.9 & $-$5.5 & $-$6.6 & $-$7.0 \\
RPCD & $-$1.9 & $-$2.8 & $-$4.0 & $-$2.9 \\
ID & $-$2.1 & $-$3.0 & $-$3.7 & $-$3.1 \\
Strict CAD & $-$4.8 & $-$8.4 & $-$10.6 & $-$9.7 \\
\bottomrule
}
\caption{Supported plain answers on ordinary tasks, grouped by their closed-book probability $p(a\mid q)$. Rows give the share, plain F1 and decoder F1 changes. Test split, six models.}
\label{tab:mech}
\end{table}

%% file: arxiv/tables/abltask.tex
% generated by scripts/v8_numbers.py from review_c1/abltask.json; split=test
\begin{table*}[t]
\centering\small
\fulltable{\textwidth}{9}{lcccccccc}{%
\toprule
\hd{Variant} & \hd{NQ-Swap} & \hd{FaithEval} & \hd{NQ-Orig.} & \hd{SQuAD} & \hd{HotpotQA} & \hd{CF} & \hd{Ord.} & \hd{Mean} \\
\midrule
Sampling & 56.9 & 63.6 & 76.0 & 84.5 & 63.5 & 60.3 & 74.7 & 67.5 \\
\oursrow \aor{} & 65.2 & 80.4 & 78.2 & 87.5 & 65.2 & 72.8 & 77.0 & 74.9 \\
Repair every answer & 66.6 & 80.4 & 76.5 & 86.8 & 63.8 & 73.5 & 75.7 & 74.6 \\
No memory veto & 65.2 & 74.0 & 78.2 & 87.5 & 65.2 & 69.6 & 77.0 & 73.3 \\
CAD only & 62.8 & 79.5 & 76.9 & 86.7 & 64.5 & 71.2 & 76.0 & 73.6 \\
ID only & 64.2 & 76.5 & 78.1 & 86.9 & 65.2 & 70.4 & 76.8 & 73.6 \\
Resample & 58.1 & 75.2 & 76.8 & 85.3 & 64.1 & 66.7 & 75.4 & 71.0 \\
Resample, strict prompt & 61.2 & 74.7 & 77.4 & 85.7 & 64.5 & 67.9 & 75.8 & 71.9 \\
Plain answer as fallback & 64.5 & 67.2 & 78.5 & 87.4 & 65.2 & 65.9 & 77.0 & 71.5 \\
\bottomrule
}
\caption{Variants of \aor{} on each task, test split, six models. The plain answer as fallback returns the plain answer, instead of the first answer not rejected by the veto, when no draw is supported.}
\label{tab:abltask}
\end{table*}

%% file: arxiv/tables/faithdecomp.tex
% generated by scripts/v8_numbers.py from review_c1/faith.json; split=test
\begin{table}[t]
\centering\small
\fulltable{\columnwidth}{3}{lcc}{%
\toprule
\hd{Decoder} & \hd{Without veto} & \hd{With veto} \\
\midrule
Veto score as a solver & 61.4 & -- \\
Sampling & 63.6 & 74.5 \\
ID & 67.8 & 76.1 \\
CAD & 73.7 & 79.2 \\
\oursrow \aor{} & 74.0 & 80.4 \\
\bottomrule
}
\caption{FaithEval accuracy of each decoder without and with the memory veto, which masks the letter of a vetoed closed-book answer. The first row picks the option with the highest veto score. Test split, six models.}
\label{tab:faith}
\end{table}

%% file: arxiv/tables/vetobase.tex
% generated by scripts/v8_numbers.py from review_c2/vote_veto.json; split=test
\begin{table}[t]
\centering\small
\fulltable{\columnwidth}{4}{lccc}{%
\toprule
\hd{Method} & \hd{CF} & \hd{Ord.} & \hd{Mean} \\
\midrule
Sampling + veto & 65.7 & 74.7 & 70.2 \\
CAD + veto & 70.8 & 72.6 & 71.7 \\
AdaCAD + veto & 68.4 & \underline{74.8} & 71.6 \\
COIECD + veto & 70.8 & 72.8 & 71.8 \\
NWCAD + veto & 62.4 & \underline{74.8} & 68.6 \\
RPCD + veto & 69.9 & 74.4 & 72.2 \\
ID + veto & 70.2 & 74.7 & \underline{72.4} \\
Strict prompt + veto & 66.9 & 74.6 & 70.8 \\
Strict CAD + veto & \underline{71.6} & 71.1 & 71.3 \\
\oursrow \aor{} (ours) & \textbf{72.8} & \textbf{77.0} & \textbf{74.9} \\
\bottomrule
}
\caption{Each baseline with the memory veto of \aor{}, which on FaithEval masks the letter of a vetoed closed-book answer. Test split, six models.}
\label{tab:vetobase}
\end{table}

%% file: arxiv/tables/span.tex
% generated by scripts/v8_numbers.py from aor_table_span_r5.json and cost_span_r5.json; split=test
\begin{table}[t]
\centering\small
\fulltable{\columnwidth}{5}{lcccc}{%
\toprule
\hd{Method} & \hd{NQ-Swap} & \hd{Ord.} & \hd{Mean} & \hd{Cost} \\
\midrule
Sampling & 56.9 & 74.7 & 65.8 & 1.00 \\
Span decoding & 61.4 & \underline{76.3} & 68.8 & 1.00 \\
Copy prompt & 59.6 & 73.4 & 66.5 & 1.02 \\
Copy prompt, CAD & 59.7 & 66.2 & 63.0 & 1.20 \\
\grouprow{5}{\aor{} with one repair} \\
Feedback & 58.5 & 74.9 & 66.7 & 1.28 \\
CAD & \underline{62.8} & 76.0 & \underline{69.4} & 1.27 \\
\midrule
\oursrow \aor{} (ours) & \textbf{65.2} & \textbf{77.0} & \textbf{71.1} & 1.53 \\
\bottomrule
}
\caption{Span and copy methods and \aor{} variants on free-form tasks. Ord.\ averages NQ-Original, SQuAD and HotpotQA, and Mean averages NQ-Swap and Ord. Cost counts processed tokens relative to sampling. Test split, six models.}
\label{tab:span}
\end{table}

%% file: arxiv/tables/fivepsg.tex
% generated by scripts/v8_numbers.py from aor_table_mp_r5.json; split=test
\begin{table}[t]
\centering\small
\fulltable{\columnwidth}{4}{lccc}{%
\toprule
\hd{Method} & \hd{NQ-Swap} & \hd{NQ-Orig.} & \hd{Mean} \\
\midrule
Sampling & 53.8 & \underline{72.7} & 63.3 \\
CAD & 58.8 & 68.5 & 63.6 \\
AdaCAD & 57.9 & 71.9 & 64.9 \\
COIECD & 58.4 & 68.6 & 63.5 \\
NWCAD & 54.3 & \underline{72.7} & 63.5 \\
RPCD & 59.1 & 72.6 & 65.8 \\
ID & \underline{60.7} & 71.8 & \underline{66.2} \\
Strict prompt & 56.0 & 71.8 & 63.9 \\
Strict CAD & 59.6 & 65.4 & 62.5 \\
\oursrow \aor{} (ours) & \textbf{61.2} & \textbf{75.5} & \textbf{68.4} \\
\bottomrule
}
\caption{NQ-Swap and NQ-Original with the passage of each question placed among four passages of other questions, at a random position. Test split, six models.}
\label{tab:fivepsg}
\end{table}

%% file: arxiv/tables/contain.tex
% generated by scripts/v8_numbers.py from review_c1/contain.json; split=test
\begin{table}[t]
\centering\small
\fulltable{\columnwidth}{4}{lccc}{%
\toprule
\hd{Method} & \hd{CF} & \hd{Ord.} & \hd{Mean} \\
\midrule
Sampling & 64.2 & 76.6 & 70.4 \\
CAD & 72.8 & 75.5 & \underline{74.1} \\
AdaCAD & 67.4 & \underline{76.8} & 72.1 \\
COIECD & 72.7 & 75.6 & \underline{74.1} \\
NWCAD & 64.6 & 76.6 & 70.6 \\
RPCD & 70.7 & 76.5 & 73.6 \\
ID & 68.8 & 74.0 & 71.4 \\
Strict prompt & 66.2 & 76.1 & 71.1 \\
Strict CAD & \underline{74.1} & 74.0 & \underline{74.1} \\
\oursrow \aor{} (ours) & \textbf{75.8} & \textbf{77.2} & \textbf{76.5} \\
\bottomrule
}
\caption{Scores when a free-form answer counts as correct if it contains a gold answer as whole words, which does not penalize longer answers. FaithEval keeps accuracy. Test split, six models.}
\label{tab:contain}
\end{table}

%% file: arxiv/tables/judge.tex
% generated by scripts/v8_numbers.py from review_c3/judge.json; split=test
\begin{table}[t]
\centering\fontsize{8.1}{9.3}\selectfont
\fulltable{\columnwidth}{6}{lccccc}{%
\toprule
\hd{Method} & \hd{NQ-Orig.} & \hd{SQuAD} & \hd{HotpotQA} & \hd{Ord.} & \hd{NQ-Swap} \\
\midrule
CAD & $-$4.6 & $-$1.8 & $-$4.9 & $-$3.8 & +0.9 \\
AdaCAD & $-$0.6 & \underline{0.0} & $-$1.2 & $-$0.6 & +1.8 \\
COIECD & $-$4.2 & $-$1.8 & $-$4.5 & $-$3.5 & +1.0 \\
NWCAD & \underline{$-$0.1} & $-$0.2 & \underline{0.0} & \underline{$-$0.1} & 0.0 \\
RPCD & $-$2.3 & $-$1.7 & $-$3.0 & $-$2.3 & +0.8 \\
ID & $-$3.4 & $-$1.0 & $-$1.0 & $-$1.8 & \underline{+2.8} \\
Strict prompt & $-$0.8 & $-$0.3 & $-$0.9 & $-$0.7 & +1.7 \\
Strict CAD & $-$7.1 & $-$3.3 & $-$7.9 & $-$6.1 & +2.0 \\
\oursrow \aor{} (ours) & \textbf{0.0} & \textbf{+0.8} & \textbf{+0.2} & \textbf{+0.3} & \textbf{+3.9} \\
\bottomrule
}
\caption{Accuracy differences from sampling on free-form answers, judged by Qwen2.5-32B-Instruct. Test split, six models.}
\label{tab:judge}
\end{table}

%% file: arxiv/tables/sent.tex
% generated by scripts/v8_numbers.py from aor_table_sent_r5.json; split=test
\begin{table}[t]
\centering\small
\fulltable{\columnwidth}{6}{lccccc}{%
\toprule
\hd{Method} & \hd{NQ-Swap} & \hd{NQ-Orig.} & \hd{SQuAD} & \hd{Ord.} & \hd{Mean} \\
\midrule
Sampling & 64.9 & 86.3 & 90.9 & 88.6 & 76.7 \\
\oursrow \aor{} (ours) & 73.3 & 86.7 & 91.9 & 89.3 & 81.3 \\
\midrule
Difference & +8.4$^{*}$ & +0.4 & +1.0 & +0.7$^{*}$ & +4.6$^{*}$ \\
\bottomrule
}
\caption{Answers given as one sentence, correct when they contain a gold answer. Ord.\ averages NQ-Original and SQuAD, and Mean averages NQ-Swap and Ord. $^{*}$: the 95\% interval of the difference excludes zero. Test split, six models.}
\label{tab:sent}
\end{table}

%% file: arxiv/tables/greedy.tex
% generated by scripts/v8_numbers.py from aor_table_grdrt_r5.json; split=test
\begin{table}[t]
\centering\small
\fulltable{\columnwidth}{8}{lccccccc}{%
\toprule
 & & & & \multicolumn{4}{c}{\hd{Mean}} \\
\cmidrule(lr){5-8}
\hd{Method} & \hd{CF} & \hd{Ord.} & \hd{Mean} & \hd{Llama} & \hd{Mistral} & \hd{OLMo} & \hd{Yi} \\
\midrule
Greedy & 58.4 & \underline{74.8} & 66.6 & 70.2 & 58.0 & 69.4 & \underline{68.8} \\
CAD & 63.3 & 71.4 & 67.3 & 74.1 & 62.1 & 67.7 & 65.4 \\
AdaCAD & 61.7 & 73.5 & 67.6 & 72.3 & 61.1 & 69.0 & 68.1 \\
COIECD & 60.8 & 72.2 & 66.5 & 72.5 & 60.5 & 67.8 & 65.3 \\
NWCAD & 58.5 & 74.7 & 66.6 & 70.5 & 58.1 & 69.3 & 68.3 \\
RPCD & 62.5 & 73.0 & 67.8 & 72.6 & \underline{63.8} & 68.5 & 66.2 \\
ID & 61.5 & 73.8 & 67.6 & 73.9 & 63.2 & 68.2 & 65.2 \\
Strict prompt & 61.9 & 74.7 & 68.3 & 74.1 & 60.0 & \underline{71.8} & 67.4 \\
Strict CAD & \underline{66.7} & 71.4 & \underline{69.1} & \underline{75.4} & \underline{63.8} & 68.8 & 68.2 \\
\oursrow \aor{} (ours) & \textbf{70.8} & \textbf{77.1} & \textbf{73.9} & \textbf{77.4} & \textbf{66.6} & \textbf{76.4} & \textbf{75.3} \\
\bottomrule
}
\caption{Greedy decoding of every method on Llama-3.1-8B, Mistral-7B, OLMo-2-13B and Yi-1.5-9B, test split, with every setting chosen again under greedy decoding on the development split. The memory veto of \aor{} uses the sampled closed-book answer of the main runs.}
\label{tab:greedy}
\end{table}

%% file: arxiv/tables/permodel.tex
% generated by scripts/v8_numbers.py from aor_table_test_r5.json; split=test
\begin{table*}[!t]
\centering\small
\fulltable{\textwidth}{7}{lcccccc}{%
\toprule
\hd{Method} & \hd{Qwen3-1.7B} & \hd{Qwen3-4B} & \hd{Qwen3-8B} & \hd{Qwen3-14B} & \hd{Llama-3.1-8B} & \hd{Mistral-7B} \\
\midrule
Sampling & 65.1 & 68.1 & 72.2 & 72.2 & 69.5 & 57.7 \\
CAD & 65.3 & 70.6 & \underline{76.2} & 75.5 & 72.3 & 62.3 \\
AdaCAD & \underline{65.9} & 69.4 & 73.3 & 73.2 & 71.9 & 60.9 \\
COIECD & 65.5 & 70.6 & 76.1 & 75.5 & 72.4 & 62.2 \\
NWCAD & 65.4 & 68.3 & 72.3 & 72.2 & 69.8 & 57.9 \\
RPCD & 65.8 & 70.4 & 74.3 & 74.6 & 71.2 & \underline{65.1} \\
ID & 62.9 & \underline{71.9} & 75.5 & 75.4 & 71.4 & \underline{65.1} \\
Strict prompt & 64.1 & 68.5 & 73.0 & 73.0 & \underline{73.3} & 59.6 \\
Strict CAD & 64.1 & 70.5 & 76.1 & \underline{75.8} & 72.7 & 63.9 \\
\oursrow \aor{} (ours) & \textbf{69.0} & \textbf{74.7} & \textbf{79.0} & \textbf{80.3} & \textbf{77.7} & \textbf{68.6} \\
\bottomrule
}
\caption{Mean of the counterfactual and ordinary averages for each model in the main comparison. Test split.}
\label{tab:permodel}
\end{table*}

%% file: main_arxiv.bbl
\begin{thebibliography}{67}
\providecommand{\natexlab}[1]{#1}

\bibitem[{{01.AI}(2024)}]{01ai2024yi15chat}
{01.AI}. 2024.
\newblock \href {https://huggingface.co/01-ai/Yi-1.5-9B-Chat}
  {{Yi-1.5-9B-Chat}}.
\newblock Hugging Face model card.

\bibitem[{Asai et~al.(2024)Asai, Wu, Wang, Sil, and
  Hajishirzi}]{asai2024selfrag}
Akari Asai, Zeqiu Wu, Yizhong Wang, Avirup Sil, and Hannaneh Hajishirzi. 2024.
\newblock \href {https://openreview.net/forum?id=hSyW5go0v8} {Self-{RAG}:
  Learning to retrieve, generate, and critique through self-reflection}.
\newblock In \emph{The Twelfth International Conference on Learning
  Representations}.

\bibitem[{Berg-Kirkpatrick et~al.(2012)Berg-Kirkpatrick, Burkett, and
  Klein}]{berg-kirkpatrick-etal-2012-empirical}
Taylor Berg-Kirkpatrick, David Burkett, and Dan Klein. 2012.
\newblock An empirical investigation of statistical significance in {NLP}.
\newblock In \emph{Proceedings of the 2012 Joint Conference on Empirical
  Methods in Natural Language Processing and Computational Natural Language
  Learning}, pages 995--1005, Jeju Island, Korea. Association for Computational
  Linguistics.

\bibitem[{Bi et~al.(2025)Bi, Huang, Wang, Yang, Zhang, Huang, Mei, Fang, Li,
  Wei, Deng, Sun, Zhang, and Liu}]{bi2025context}
Baolong Bi, Shaohan Huang, Yiwei Wang, Tianchi Yang, Zihan Zhang, Haizhen
  Huang, Lingrui Mei, Junfeng Fang, Zehao Li, Furu Wei, Weiwei Deng, Feng Sun,
  Qi~Zhang, and Shenghua Liu. 2025.
\newblock \href {https://doi.org/10.18653/v1/2025.findings-acl.536}
  {{Context-DPO: Aligning Language Models for Context-Faithfulness}}.
\newblock In \emph{{Findings of the Association for Computational Linguistics:
  ACL 2025}}.

\bibitem[{Chen et~al.(2024)Chen, Zaharia, and Zou}]{chen2024frugalgpt}
Lingjiao Chen, Matei Zaharia, and James Zou. 2024.
\newblock \href {https://openreview.net/forum?id=cSimKw5p6R} {{FrugalGPT}: How
  to use large language models while reducing cost and improving performance}.
\newblock \emph{Transactions on Machine Learning Research}.

\bibitem[{Chuang et~al.(2024)Chuang, Qiu, Hsieh, Krishna, Kim, and
  Glass}]{chuang-etal-2024-lookback}
Yung-Sung Chuang, Linlu Qiu, Cheng-Yu Hsieh, Ranjay Krishna, Yoon Kim, and
  James~R. Glass. 2024.
\newblock Lookback lens: Detecting and mitigating contextual hallucinations in
  large language models using only attention maps.
\newblock In \emph{Proceedings of the 2024 Conference on Empirical Methods in
  Natural Language Processing}, pages 1419--1436, Miami, Florida, USA.
  Association for Computational Linguistics.

\bibitem[{Cobbe et~al.(2021)Cobbe, Kosaraju, Bavarian, Chen, Jun, Kaiser,
  Plappert, Tworek, Hilton, Nakano, Hesse, and Schulman}]{cobbe2021gsm8k}
Karl Cobbe, Vineet Kosaraju, Mohammad Bavarian, Mark Chen, Heewoo Jun, Lukasz
  Kaiser, Matthias Plappert, Jerry Tworek, Jacob Hilton, Reiichiro Nakano,
  Christopher Hesse, and John Schulman. 2021.
\newblock Training verifiers to solve math word problems.
\newblock \emph{arXiv preprint arXiv:2110.14168}.

\bibitem[{Dhuliawala et~al.(2024)Dhuliawala, Komeili, Xu, Raileanu, Li,
  Celikyilmaz, and Weston}]{dhuliawala-etal-2024-chain}
Shehzaad Dhuliawala, Mojtaba Komeili, Jing Xu, Roberta Raileanu, Xian Li, Asli
  Celikyilmaz, and Jason Weston. 2024.
\newblock \href {https://doi.org/10.18653/v1/2024.findings-acl.212}
  {Chain-of-verification reduces hallucination in large language models}.
\newblock In \emph{Findings of the Association for Computational Linguistics:
  ACL 2024}.

\bibitem[{Fisch et~al.(2019)Fisch, Talmor, Jia, Seo, Choi, and
  Chen}]{fisch-etal-2019-mrqa}
Adam Fisch, Alon Talmor, Robin Jia, Minjoon Seo, Eunsol Choi, and Danqi Chen.
  2019.
\newblock \href {https://doi.org/10.18653/v1/D19-5801} {{MRQA} 2019 shared
  task: Evaluating generalization in reading comprehension}.
\newblock In \emph{Proceedings of the 2nd Workshop on Machine Reading for
  Question Answering}.

\bibitem[{Gao et~al.(2023)Gao, Dai, Pasupat, Chen, Chaganty, Fan, Zhao, Lao,
  Lee, Juan, and Guu}]{gao-etal-2023-rarr}
Luyu Gao, Zhuyun Dai, Panupong Pasupat, Anthony Chen, Arun~Tejasvi Chaganty,
  Yicheng Fan, Vincent Zhao, Ni~Lao, Hongrae Lee, Da-Cheng Juan, and Kelvin
  Guu. 2023.
\newblock \href {https://doi.org/10.18653/v1/2023.acl-long.910} {{RARR}:
  Researching and revising what language models say, using language models}.
\newblock In \emph{Proceedings of the 61st Annual Meeting of the Association
  for Computational Linguistics (Volume 1: Long Papers)}.

\bibitem[{Gema et~al.(2025)Gema, Jin, Abdulaal, Diethe, Teare, Alex, Minervini,
  and Saseendran}]{gema-etal-2025-decore}
Aryo~Pradipta Gema, Chen Jin, Ahmed Abdulaal, Tom Diethe, Philip~Alexander
  Teare, Beatrice Alex, Pasquale Minervini, and Amrutha Saseendran. 2025.
\newblock {D}e{C}o{R}e: Decoding by contrasting retrieval heads to mitigate
  hallucinations.
\newblock In \emph{Findings of the Association for Computational Linguistics:
  EMNLP 2025}, pages 10003--10039, Suzhou, China. Association for Computational
  Linguistics.

\bibitem[{{Gemma Team}(2026)}]{gemmateam2026gemma4}
{Gemma Team}. 2026.
\newblock \href {https://arxiv.org/abs/2607.02770} {Gemma 4 technical report}.
\newblock \emph{arXiv preprint arXiv:2607.02770}.

\bibitem[{He et~al.(2025)He, Pandey, Schrum, and Dragan}]{he2025cos}
Jerry Zhi-Yang He, Sashrika Pandey, Mariah~L. Schrum, and Anca Dragan. 2025.
\newblock Context steering: Controllable personalization at inference time.
\newblock In \emph{The Thirteenth International Conference on Learning
  Representations}.

\bibitem[{He et~al.(2023)He, Gao, and Chen}]{he2023debertav3}
Pengcheng He, Jianfeng Gao, and Weizhu Chen. 2023.
\newblock \href {https://openreview.net/forum?id=sE7-XhLxHA} {{DeBERTaV3:
  Improving DeBERTa using ELECTRA-Style Pre-Training with Gradient-Disentangled
  Embedding Sharing}}.
\newblock In \emph{{The Eleventh International Conference on Learning
  Representations}}.

\bibitem[{Holtzman et~al.(2020)Holtzman, Buys, Du, Forbes, and
  Choi}]{holtzman2020curious}
Ari Holtzman, Jan Buys, Li~Du, Maxwell Forbes, and Yejin Choi. 2020.
\newblock The curious case of neural text degeneration.
\newblock In \emph{International Conference on Learning Representations}.

\bibitem[{Jiang et~al.(2023)Jiang, Sablayrolles, Mensch, Bamford, Chaplot,
  de~las Casas, Bressand, Lengyel, Lample, Saulnier, Lavaud, Lachaux, Stock,
  Le~Scao, Lavril, Wang, Lacroix, and El~Sayed}]{jiang2023mistral}
Albert~Q. Jiang, Alexandre Sablayrolles, Arthur Mensch, Chris Bamford,
  Devendra~Singh Chaplot, Diego de~las Casas, Florian Bressand, Gianna Lengyel,
  Guillaume Lample, Lucile Saulnier, L{\'e}lio~Renard Lavaud, Marie-Anne
  Lachaux, Pierre Stock, Teven Le~Scao, Thibaut Lavril, Thomas Wang,
  Timoth{\'e}e Lacroix, and William El~Sayed. 2023.
\newblock Mistral {7B}.
\newblock \emph{arXiv preprint arXiv:2310.06825}.

\bibitem[{Jiang et~al.(2026)Jiang, Wu, Wang, Zhu, and
  Huang}]{jiang2026conflictaware}
Runze Jiang, Taiqiang Wu, Yan Wang, Bingyu Zhu, and Longtao Huang. 2026.
\newblock From context-aware to conflict-aware: Generalizing contrastive
  decoding for knowledge conflict in {LLM}s.
\newblock \emph{arXiv preprint arXiv:2606.10298}.

\bibitem[{Joshi et~al.(2017)Joshi, Choi, Weld, and
  Zettlemoyer}]{joshi-etal-2017-triviaqa}
Mandar Joshi, Eunsol Choi, Daniel Weld, and Luke Zettlemoyer. 2017.
\newblock {T}rivia{QA}: A large scale distantly supervised challenge dataset
  for reading comprehension.
\newblock In \emph{Proceedings of the 55th Annual Meeting of the Association
  for Computational Linguistics (Volume 1: Long Papers)}, pages 1601--1611,
  Vancouver, Canada. Association for Computational Linguistics.

\bibitem[{Kadavath et~al.(2022)Kadavath, Conerly, Askell, Henighan, Drain,
  Perez, Schiefer, Hatfield-Dodds, DasSarma, Tran-Johnson, Johnston, El-Showk,
  Jones, Elhage, Hume, Chen, Bai, Bowman, Fort, Ganguli, Hernandez, Jacobson,
  Kernion, Kravec, Lovitt, Ndousse, Olsson, Ringer, Amodei, Brown, Clark,
  Joseph, Mann, McCandlish, Olah, and Kaplan}]{kadavath2022language}
Saurav Kadavath, Tom Conerly, Amanda Askell, Tom Henighan, Dawn Drain, Ethan
  Perez, Nicholas Schiefer, Zac Hatfield-Dodds, Nova DasSarma, Eli
  Tran-Johnson, Scott Johnston, Sheer El-Showk, Andy Jones, Nelson Elhage,
  Tristan Hume, Anna Chen, Yuntao Bai, Sam Bowman, Stanislav Fort, and 17
  others. 2022.
\newblock \href {https://arxiv.org/abs/2207.05221} {{Language Models (Mostly)
  Know What They Know}}.
\newblock \emph{{arXiv preprint}}, arXiv:2207.05221.

\bibitem[{Kang et~al.(2026)Kang, Lee, Seo, and Lim}]{kang2026thinkfuse}
Myunghoon Kang, Jungseob Lee, Jaehyung Seo, and Heuiseok Lim. 2026.
\newblock \href {https://arxiv.org/abs/2610.07803} {Thinkfuse: Trajectory-aware
  test-time fusion for small reasoning models}.
\newblock \emph{arXiv preprint}, arXiv:2610.07803.

\bibitem[{Khandelwal et~al.(2025)Khandelwal, Gupta, and
  Agrawal}]{khandelwal-etal-2025-cocoa}
Anant Khandelwal, Manish Gupta, and Puneet Agrawal. 2025.
\newblock {C}o{C}o{A}: Confidence- and context-aware adaptive decoding for
  resolving knowledge conflicts in large language models.
\newblock In \emph{Proceedings of the 2025 Conference on Empirical Methods in
  Natural Language Processing}, pages 6835--6855, Suzhou, China. Association
  for Computational Linguistics.

\bibitem[{Kim et~al.(2024{\natexlab{a}})Kim, Kim, Lee, and
  Yun}]{kim2024instructive}
Taehyeon Kim, Joonkee Kim, Gihun Lee, and Se-Young Yun. 2024{\natexlab{a}}.
\newblock Instructive decoding: Instruction-tuned large language models are
  self-refiner from noisy instructions.
\newblock In \emph{The Twelfth International Conference on Learning
  Representations}.

\bibitem[{Kim et~al.(2024{\natexlab{b}})Kim, Kim, Park, Park, Cho, Kim, Yoo,
  Lee, and Kim}]{kim-etal-2024-adaptive}
Youna Kim, Hyuhng~Joon Kim, Cheonbok Park, Choonghyun Park, Hyunsoo Cho,
  Junyeob Kim, Kang~Min Yoo, Sang-goo Lee, and Taeuk Kim. 2024{\natexlab{b}}.
\newblock Adaptive contrastive decoding in retrieval-augmented generation for
  handling noisy contexts.
\newblock In \emph{Findings of the Association for Computational Linguistics:
  EMNLP 2024}, pages 2421--2431, Miami, Florida, USA. Association for
  Computational Linguistics.

\bibitem[{Kwiatkowski et~al.(2019)Kwiatkowski, Palomaki, Redfield, Collins,
  Parikh, Alberti, Epstein, Polosukhin, Devlin, Lee, Toutanova, Jones, Kelcey,
  Chang, Dai, Uszkoreit, Le, and Petrov}]{kwiatkowski-etal-2019-natural}
Tom Kwiatkowski, Jennimaria Palomaki, Olivia Redfield, Michael Collins, Ankur
  Parikh, Chris Alberti, Danielle Epstein, Illia Polosukhin, Jacob Devlin,
  Kenton Lee, Kristina Toutanova, Llion Jones, Matthew Kelcey, Ming-Wei Chang,
  Andrew~M. Dai, Jakob Uszkoreit, Quoc Le, and Slav Petrov. 2019.
\newblock {N}atural {Q}uestions: A benchmark for question answering research.
\newblock \emph{Transactions of the Association for Computational Linguistics},
  7:452--466.

\bibitem[{Laurer et~al.(2023)Laurer, van Atteveldt, Casas, and
  Welbers}]{laurer2023less}
Moritz Laurer, Wouter van Atteveldt, Andreu Casas, and Kasper Welbers. 2023.
\newblock \href {https://doi.org/10.1017/pan.2023.20} {{Less Annotating, More
  Classifying: Addressing the Data Scarcity Issue of Supervised Machine
  Learning with Deep Transfer Learning and BERT-NLI}}.
\newblock \emph{{Political Analysis}}.

\bibitem[{Lee et~al.(2026{\natexlab{a}})Lee, Lee, Park, Moon, and
  Lim}]{lee2026pricecheck}
Dongyub~Jude Lee, Jungseob Lee, Chanjun Park, Hyeonseok Moon, and Heuiseok Lim.
  2026{\natexlab{a}}.
\newblock \href {https://arxiv.org/abs/2609.37493} {Risk-controlled selective
  llm answering by pricing label-free checks}.
\newblock \emph{arXiv preprint}, arXiv:2609.37493.

\bibitem[{Lee et~al.(2026{\natexlab{b}})Lee, Eo, Hong, Lee, Park, Seo, and
  Lim}]{lee2026directional}
Jungseob Lee, Sugyeong Eo, Seongtae Hong, Seungyoon Lee, Chanjun Park, Jaehyung
  Seo, and Heuiseok Lim. 2026{\natexlab{b}}.
\newblock \href {https://arxiv.org/abs/2610.00997} {Distilling directional
  verification}.
\newblock \emph{arXiv preprint}, arXiv:2610.00997.

\bibitem[{Lee et~al.(2026{\natexlab{c}})Lee, Hong, Lee, Seo, Son, Eo, Park,
  Park, Moon, and Lim}]{lee2026dart_routing}
Jungseob Lee, Seongtae Hong, Seungjun Lee, Jaehyung Seo, Junyoung Son, Sugyeong
  Eo, Chanjun Park, Hyeongju Park, Hyeonseok Moon, and Heuiseok Lim.
  2026{\natexlab{c}}.
\newblock \href {https://arxiv.org/abs/2606.23181} {Dart: Draft-agreement
  routing for training-free adaptive thinking budgets in hybrid reasoning
  models}.
\newblock \emph{arXiv preprint}, arXiv:2606.23181.

\bibitem[{Lee et~al.(2025)Lee, Kim, Yoon, Hong, Jang, Lee, Seo, Park, Park, and
  Lim}]{lee2025kullmrag}
Jungseob Lee, Minhyuk Kim, Jeongho Yoon, Seongtae Hong, Youngjoon Jang,
  Seungyoon Lee, Jaehyung Seo, Chanjun Park, Jeongbae Park, and Heuiseok Lim.
  2025.
\newblock \href
  {https://sites.google.com/view/hclt-2025/%ED%94%84%EB%A1%9C%EA%B7%B8%EB%9E%A8}
  {Kullm-rag: A rag-specialized large language model with intelligent
  self-verification}.
\newblock In \emph{HCLT-KACL 2025}.

\bibitem[{Lee et~al.(2026{\natexlab{d}})Lee, Lee, Park, Eo, and
  Lim}]{lee2026redline}
Jungseob Lee, Dongyub~Jude Lee, Chanjun Park, Sugyeong Eo, and Heuiseok Lim.
  2026{\natexlab{d}}.
\newblock \href {https://arxiv.org/abs/2609.33887} {Faster block-diffusion
  serving with distribution-free risk guarantees}.
\newblock \emph{arXiv preprint}, arXiv:2609.33887.

\bibitem[{Lee and Park(2026)}]{lee2026agentloop}
Jungseob Lee and Chanjun Park. 2026.
\newblock \href {https://doi.org/10.2139/ssrn.7186738} {The agent loop: A
  survey of control strategies, skills, and harnesses for llm agents}.
\newblock \emph{SSRN working paper}.

\bibitem[{Lee et~al.(2026{\natexlab{e}})Lee, Park, and Lim}]{lee2026madara}
Jungseob Lee, Chanjun Park, and Heuiseok Lim. 2026{\natexlab{e}}.
\newblock \href {https://arxiv.org/abs/2606.25191} {To isolate or to score?
  model-adaptive assessment for cost-efficient multi-agent rag}.
\newblock \emph{arXiv preprint}, arXiv:2606.25191.

\bibitem[{Lee et~al.(2026{\natexlab{f}})Lee, Seo, and Lim}]{lee2026layermix}
Jungseob Lee, Jaehyung Seo, and Heuiseok Lim. 2026{\natexlab{f}}.
\newblock \href {https://arxiv.org/abs/2608.28930} {The hallucination signal is
  a mean shift: Why simple probes suffice}.
\newblock \emph{arXiv preprint}, arXiv:2608.28930.

\bibitem[{Lee et~al.(2022)Lee, Shim, Son, Park, Kim, and Lim}]{lee2022norose}
Jungseob Lee, Midan Shim, Suhyune Son, Chanjun Park, Yujin Kim, and Heuiseok
  Lim. 2022.
\newblock \href {https://arxiv.org/abs/2201.03239} {There is no rose without a
  thorn: Finding weaknesses on blenderbot 2.0 in terms of model, data and
  user-centric approach}.
\newblock \emph{arXiv preprint}, arXiv:2201.03239.

\bibitem[{Li et~al.(2026)Li, Khondaker, Abaskohi, Murray, Carenini, and
  Laradji}]{li2026dccd}
Raymond Li, Md~Tawkat~Islam Khondaker, Amirhossein Abaskohi, Gabriel Murray,
  Giuseppe Carenini, and Issam~H. Laradji. 2026.
\newblock Dual-confidence contrastive decoding for retrieval-augmented
  generation.
\newblock \emph{arXiv preprint arXiv:2607.00570}.

\bibitem[{{Llama Team}(2024)}]{grattafiori2024llama}
{Llama Team}. 2024.
\newblock The {Llama} 3 herd of models.
\newblock \emph{arXiv preprint arXiv:2407.21783}.

\bibitem[{Longpre et~al.(2021)Longpre, Perisetla, Chen, Ramesh, DuBois, and
  Singh}]{longpre-etal-2021-entity}
Shayne Longpre, Kartik Perisetla, Anthony Chen, Nikhil Ramesh, Chris DuBois,
  and Sameer Singh. 2021.
\newblock Entity-based knowledge conflicts in question answering.
\newblock In \emph{Proceedings of the 2021 Conference on Empirical Methods in
  Natural Language Processing}, pages 7052--7063, Online and Punta Cana,
  Dominican Republic. Association for Computational Linguistics.

\bibitem[{Minder et~al.(2025)Minder, Du, Stoehr, Monea, Wendler, West, and
  Cotterell}]{minder2025controllable}
Julian Minder, Kevin Du, Niklas Stoehr, Giovanni Monea, Chris Wendler, Robert
  West, and Ryan Cotterell. 2025.
\newblock Controllable context sensitivity and the knob behind it.
\newblock In \emph{The Thirteenth International Conference on Learning
  Representations}.

\bibitem[{Ming et~al.(2025)Ming, Purushwalkam, Pandit, Ke, Nguyen, Xiong, and
  Joty}]{ming2025faitheval}
Yifei Ming, Senthil Purushwalkam, Shrey Pandit, Zixuan Ke, Xuan-Phi Nguyen,
  Caiming Xiong, and Shafiq Joty. 2025.
\newblock {FaithEval}: Can your language model stay faithful to context, even
  if ``the moon is made of marshmallows''.
\newblock In \emph{International Conference on Learning Representations}.

\bibitem[{{Mistral AI}(2024)}]{mistral2024instructv03}
{Mistral AI}. 2024.
\newblock \href {https://huggingface.co/mistralai/Mistral-7B-Instruct-v0.3}
  {{Mistral-7B-Instruct-v0.3}}.
\newblock Hugging Face model card.

\bibitem[{Paszke et~al.(2019)Paszke, Gross, Massa, Lerer, Bradbury, Chanan,
  Killeen, Lin, Gimelshein, Antiga, Desmaison, Kopf, Yang, DeVito, Raison,
  Tejani, Chilamkurthy, Steiner, Fang, Bai, and Chintala}]{paszke2019pytorch}
Adam Paszke, Sam Gross, Francisco Massa, Adam Lerer, James Bradbury, Gregory
  Chanan, Trevor Killeen, Zeming Lin, Natalia Gimelshein, Luca Antiga, Alban
  Desmaison, Andreas Kopf, Edward Yang, Zachary DeVito, Martin Raison, Alykhan
  Tejani, Sasank Chilamkurthy, Benoit Steiner, Lu~Fang, and 2 others. 2019.
\newblock {PyTorch}: An imperative style, high-performance deep learning
  library.
\newblock In \emph{Advances in Neural Information Processing Systems},
  volume~32. Curran Associates, Inc.

\bibitem[{Qiu et~al.(2025)Qiu, Ou, Wu, Li, Liu, and
  King}]{qiu-etal-2025-entropy}
Zexuan Qiu, Zijing Ou, Bin Wu, Jingjing Li, Aiwei Liu, and Irwin King. 2025.
\newblock Entropy-based decoding for retrieval-augmented large language models.
\newblock In \emph{Proceedings of the 2025 Conference of the Nations of the
  Americas Chapter of the Association for Computational Linguistics: Human
  Language Technologies (Volume 1: Long Papers)}, pages 4616--4627,
  Albuquerque, New Mexico. Association for Computational Linguistics.

\bibitem[{{Qwen Team}(2024)}]{yang2024qwen2}
{Qwen Team}. 2024.
\newblock Qwen2.5 technical report.
\newblock \emph{arXiv preprint arXiv:2412.15115}.

\bibitem[{{Qwen Team}(2025)}]{yang2025qwen3}
{Qwen Team}. 2025.
\newblock Qwen3 technical report.
\newblock \emph{arXiv preprint arXiv:2505.09388}.

\bibitem[{Rajpurkar et~al.(2016)Rajpurkar, Zhang, Lopyrev, and
  Liang}]{rajpurkar-etal-2016-squad}
Pranav Rajpurkar, Jian Zhang, Konstantin Lopyrev, and Percy Liang. 2016.
\newblock {SQ}u{AD}: 100,000+ questions for machine comprehension of text.
\newblock In \emph{Proceedings of the 2016 Conference on Empirical Methods in
  Natural Language Processing}, pages 2383--2392, Austin, Texas. Association
  for Computational Linguistics.

\bibitem[{Ram et~al.(2023)Ram, Levine, Dalmedigos, Muhlgay, Shashua,
  Leyton-Brown, and Shoham}]{ram-etal-2023-context}
Ori Ram, Yoav Levine, Itay Dalmedigos, Dor Muhlgay, Amnon Shashua, Kevin
  Leyton-Brown, and Yoav Shoham. 2023.
\newblock \href {https://doi.org/10.1162/tacl_a_00605} {In-context
  retrieval-augmented language models}.
\newblock \emph{Transactions of the Association for Computational Linguistics},
  11:1316--1331.

\bibitem[{Shi et~al.(2023)Shi, Chen, Misra, Scales, Dohan, Chi, Sch{\"a}rli,
  and Zhou}]{shi2023distracted}
Freda Shi, Xinyun Chen, Kanishka Misra, Nathan Scales, David Dohan, Ed~H. Chi,
  Nathanael Sch{\"a}rli, and Denny Zhou. 2023.
\newblock Large language models can be easily distracted by irrelevant context.
\newblock In \emph{Proceedings of the 40th International Conference on Machine
  Learning}, volume 202 of \emph{Proceedings of Machine Learning Research},
  pages 31210--31227. PMLR.

\bibitem[{Shi et~al.(2024)Shi, Han, Lewis, Tsvetkov, Zettlemoyer, and
  Yih}]{shi-etal-2024-trusting}
Weijia Shi, Xiaochuang Han, Mike Lewis, Yulia Tsvetkov, Luke Zettlemoyer, and
  Wen-tau Yih. 2024.
\newblock Trusting your evidence: Hallucinate less with context-aware decoding.
\newblock In \emph{Proceedings of the 2024 Conference of the North American
  Chapter of the Association for Computational Linguistics: Human Language
  Technologies (Volume 2: Short Papers)}, pages 783--791, Mexico City, Mexico.
  Association for Computational Linguistics.

\bibitem[{Tao and Agrawal(2026)}]{tao2026noworse}
Yufei Tao and Ameeta Agrawal. 2026.
\newblock No-worse context-aware decoding: Preventing neutral regression in
  context-conditioned generation.
\newblock \emph{arXiv preprint arXiv:2604.16686}.

\bibitem[{{Team OLMo} et~al.(2024){Team OLMo}, Walsh, Soldaini, Groeneveld, Lo,
  Arora et~al.}]{olmo2024olmo2}
{Team OLMo}, Pete Walsh, Luca Soldaini, Dirk Groeneveld, Kyle Lo, Shane Arora,
  and 1 others. 2024.
\newblock 2 {OLMo} 2 {Furious}.
\newblock \emph{arXiv preprint arXiv:2501.00656}.

\bibitem[{van~der Poel et~al.(2022)van~der Poel, Cotterell, and
  Meister}]{van-der-poel-etal-2022-mutual}
Liam van~der Poel, Ryan Cotterell, and Clara Meister. 2022.
\newblock Mutual information alleviates hallucinations in abstractive
  summarization.
\newblock In \emph{Proceedings of the 2022 Conference on Empirical Methods in
  Natural Language Processing}, pages 5956--5965, Abu Dhabi, United Arab
  Emirates. Association for Computational Linguistics.

\bibitem[{Wang et~al.(2025{\natexlab{a}})Wang, Prasad, Stengel-Eskin, and
  Bansal}]{wang-etal-2025-adacad}
Han Wang, Archiki Prasad, Elias Stengel-Eskin, and Mohit Bansal.
  2025{\natexlab{a}}.
\newblock {A}da{CAD}: Adaptively decoding to balance conflicts between
  contextual and parametric knowledge.
\newblock In \emph{Proceedings of the 2025 Conference of the Nations of the
  Americas Chapter of the Association for Computational Linguistics: Human
  Language Technologies (Volume 1: Long Papers)}, pages 11636--11652,
  Albuquerque, New Mexico. Association for Computational Linguistics.

\bibitem[{Wang et~al.(2023)Wang, Wei, Schuurmans, Le, Chi, Narang, Chowdhery,
  and Zhou}]{wang2023selfconsistency}
Xuezhi Wang, Jason Wei, Dale Schuurmans, Quoc~V. Le, Ed~H. Chi, Sharan Narang,
  Aakanksha Chowdhery, and Denny Zhou. 2023.
\newblock \href {https://openreview.net/forum?id=1PL1NIMMrw} {Self-consistency
  improves chain of thought reasoning in language models}.
\newblock In \emph{The Eleventh International Conference on Learning
  Representations}.

\bibitem[{Wang et~al.(2025{\natexlab{b}})Wang, Wang, Bai, and
  Luo}]{wang-etal-2025-continuously}
Yilin Wang, Heng Wang, Yuyang Bai, and Minnan Luo. 2025{\natexlab{b}}.
\newblock Continuously steering {LLM}s sensitivity to contextual knowledge with
  proxy models.
\newblock In \emph{Proceedings of the 2025 Conference on Empirical Methods in
  Natural Language Processing}, pages 4682--4698, Suzhou, China. Association
  for Computational Linguistics.

\bibitem[{Welbl et~al.(2017)Welbl, Liu, and
  Gardner}]{welbl-etal-2017-crowdsourcing}
Johannes Welbl, Nelson~F. Liu, and Matt Gardner. 2017.
\newblock \href {https://doi.org/10.18653/v1/W17-4413} {Crowdsourcing multiple
  choice science questions}.
\newblock In \emph{Proceedings of the 3rd Workshop on Noisy User-generated
  Text}.

\bibitem[{Wolf et~al.(2020)Wolf, Debut, Sanh, Chaumond, Delangue, Moi, Cistac,
  Rault, Louf, Funtowicz, Davison, Shleifer, von Platen, Ma, Jernite, Plu, Xu,
  Le~Scao, Gugger, Drame, Lhoest, and Rush}]{wolf-etal-2020-transformers}
Thomas Wolf, Lysandre Debut, Victor Sanh, Julien Chaumond, Clement Delangue,
  Anthony Moi, Pierric Cistac, Tim Rault, Remi Louf, Morgan Funtowicz, Joe
  Davison, Sam Shleifer, Patrick von Platen, Clara Ma, Yacine Jernite, Julien
  Plu, Canwen Xu, Teven Le~Scao, Sylvain Gugger, and 3 others. 2020.
\newblock Transformers: State-of-the-art natural language processing.
\newblock In \emph{Proceedings of the 2020 Conference on Empirical Methods in
  Natural Language Processing: System Demonstrations}, pages 38--45, Online.
  Association for Computational Linguistics.

\bibitem[{Xie et~al.(2024)Xie, Zhang, Chen, Lou, and Su}]{xie2024chameleon}
Jian Xie, Kai Zhang, Jiangjie Chen, Renze Lou, and Yu~Su. 2024.
\newblock Adaptive chameleon or stubborn sloth: Revealing the behavior of large
  language models in knowledge conflicts.
\newblock In \emph{The Twelfth International Conference on Learning
  Representations}.

\bibitem[{Xu et~al.(2024)Xu, Qi, Guo, Wang, Wang, Zhang, and
  Xu}]{xu-etal-2024-knowledge-conflicts}
Rongwu Xu, Zehan Qi, Zhijiang Guo, Cunxiang Wang, Hongru Wang, Yue Zhang, and
  Wei Xu. 2024.
\newblock \href {https://doi.org/10.18653/v1/2024.emnlp-main.486} {Knowledge
  conflicts for {LLM}s: A survey}.
\newblock In \emph{Proceedings of the 2024 Conference on Empirical Methods in
  Natural Language Processing}, pages 8541--8565, Miami, Florida, USA.
  Association for Computational Linguistics.

\bibitem[{Yang et~al.(2018)Yang, Qi, Zhang, Bengio, Cohen, Salakhutdinov, and
  Manning}]{yang-etal-2018-hotpotqa}
Zhilin Yang, Peng Qi, Saizheng Zhang, Yoshua Bengio, William Cohen, Ruslan
  Salakhutdinov, and Christopher~D. Manning. 2018.
\newblock {H}otpot{QA}: A dataset for diverse, explainable multi-hop question
  answering.
\newblock In \emph{Proceedings of the 2018 Conference on Empirical Methods in
  Natural Language Processing}, pages 2369--2380, Brussels, Belgium.
  Association for Computational Linguistics.

\bibitem[{Young et~al.(2024)Young, Chen, Li, Huang et~al.}]{young2024yi}
Alex Young, Bei Chen, Chao Li, Chengen Huang, and 1 others. 2024.
\newblock Yi: Open foundation models by {01.AI}.
\newblock \emph{arXiv preprint arXiv:2403.04652}.

\bibitem[{Yuan et~al.(2024)Yuan, Yang, Wang, Liu, Zhao, and
  Liu}]{yuan-etal-2024-discerning}
Xiaowei Yuan, Zhao Yang, Yequan Wang, Shengping Liu, Jun Zhao, and Kang Liu.
  2024.
\newblock Discerning and resolving knowledge conflicts through adaptive
  decoding with contextual information-entropy constraint.
\newblock In \emph{Findings of the Association for Computational Linguistics:
  ACL 2024}, pages 3903--3922, Bangkok, Thailand. Association for Computational
  Linguistics.

\bibitem[{Zhang et~al.(2026)Zhang, Ye, Gao, Li, Wu, Tian, Duan, Du, Li, and
  Liu}]{zhang2026cfb}
Weixu Zhang, Fanghua Ye, Qiang Gao, Jian Li, Haolun Wu, Yuxing Tian, Sijing
  Duan, Nan Du, Xiaolong Li, and Xue Liu. 2026.
\newblock Context-fidelity boosting: Enhancing faithful generation through
  watermark-inspired decoding.
\newblock \emph{arXiv preprint arXiv:2604.22335}.

\bibitem[{Zhao et~al.(2025)Zhao, Devoto, Hong, Du, Gema, Wang, He, Wong, and
  Minervini}]{zhao-etal-2025-steering}
Yu~Zhao, Alessio Devoto, Giwon Hong, Xiaotang Du, Aryo~Pradipta Gema, Hongru
  Wang, Xuanli He, Kam-Fai Wong, and Pasquale Minervini. 2025.
\newblock Steering knowledge selection behaviours in {LLM}s via {SAE}-based
  representation engineering.
\newblock In \emph{Proceedings of the 2025 Conference of the Nations of the
  Americas Chapter of the Association for Computational Linguistics: Human
  Language Technologies (Volume 1: Long Papers)}, pages 5117--5136,
  Albuquerque, New Mexico. Association for Computational Linguistics.

\bibitem[{Zhao et~al.(2024)Zhao, Monti, Lehmann, and
  Assem}]{zhao-etal-2024-enhancing}
Zheng Zhao, Emilio Monti, Jens Lehmann, and Haytham Assem. 2024.
\newblock Enhancing contextual understanding in large language models through
  contrastive decoding.
\newblock In \emph{Proceedings of the 2024 Conference of the North American
  Chapter of the Association for Computational Linguistics: Human Language
  Technologies (Volume 1: Long Papers)}, pages 4225--4237, Mexico City, Mexico.
  Association for Computational Linguistics.

\bibitem[{Zhou et~al.(2023)Zhou, Zhang, Poon, and
  Chen}]{zhou-etal-2023-context}
Wenxuan Zhou, Sheng Zhang, Hoifung Poon, and Muhao Chen. 2023.
\newblock \href {https://doi.org/10.18653/v1/2023.findings-emnlp.968}
  {Context-faithful prompting for large language models}.
\newblock In \emph{Findings of the Association for Computational Linguistics:
  EMNLP 2023}, pages 14544--14556, Singapore. Association for Computational
  Linguistics.

\bibitem[{Zhou et~al.(2026)Zhou, Li, Lu, Wang, Liu, Wang, Yu, Zhang, and
  Li}]{zhou2026dcrd}
Yigeng Zhou, Wu~Li, Yifan Lu, Yequan Wang, Xuebo Liu, Wenya Wang, Jun Yu, Min
  Zhang, and Jing Li. 2026.
\newblock Mitigating context-memory conflicts in {LLM}s through dynamic
  cognitive reconciliation decoding.
\newblock \emph{IEEE Transactions on Audio, Speech and Language Processing}.
\newblock ArXiv preprint arXiv:2605.12185.

\bibitem[{Zyskind et~al.(2024)Zyskind, South, and
  Pentland}]{zyskind-etal-2024-dont}
Guy Zyskind, Tobin South, and Alex Pentland. 2024.
\newblock \href {https://aclanthology.org/2024.privatenlp-1.2/} {Don{'}t forget
  private retrieval: distributed private similarity search for large language
  models}.
\newblock In \emph{Proceedings of the Fifth Workshop on Privacy in Natural
  Language Processing}, pages 7--19, Bangkok, Thailand. Association for
  Computational Linguistics.

\end{thebibliography}
